\documentclass[Final,journal]{IEEEtran}
\usepackage{cite}
 \usepackage{graphics}
 \usepackage{graphicx}
\usepackage{array}
\usepackage{amsfonts}
\usepackage{amsmath}
\usepackage{mathrsfs}
\usepackage{multirow}

\usepackage{colortbl}
\definecolor{mygray}{gray}{0.95}
\usepackage{color}
\newcommand\ChangeRT[1]{\noalign{\hrule height #1}}

\begin{document}
\title{Person Re-identification with Metric Learning using Privileged Information}
\author{Xun~Yang,
        Meng~Wang, ~\IEEEmembership{Member,~IEEE},
          Dacheng~Tao,~\IEEEmembership{Fellow,~IEEE}
          
\thanks{©20XX IEEE. Personal use of this material is permitted. Permission from IEEE must be obtained for all other uses, in any current or future media, including reprinting/republishing this material for advertising or promotional purposes, creating new collective works, for resale or redistribution to servers or lists, or reuse of any copyrighted component of this work in other works.
}  
\thanks{This work is supported by the National 973 Program of China under grant 2014CB347600 and the National Nature Science Foundation of China under grant 61432019.}  
\thanks{Xun Yang, and Meng Wang are with the School
of Computer and Information Engineering, Hefei University of Technology, Hefei, P. R. China.
 (E-mail: hfutyangxun@gmail.com, eric.mengwang@gmail.com.)}
 \thanks{Dacheng Tao is with the UBTech Sydney Artificial Intelligence Institute and the School of Information Technologies, Faculty of Engineering and Information Technologies, The University of Sydney, Darlington, NSW 2008, Australia. (E-mail: dacheng.tao@sydney.edu.au.)
}}

\markboth{IEEE Transaction on Image Processing}%
{Shell \MakeLowercase{\textit{et al.}}: Bare Demo of IEEEtran.cls for Computer Society Journals}
\maketitle

\begin{abstract}
Despite the promising progress made in recent years, person re-identification remains a challenging task due to complex variations in human appearances from different camera views. This paper presents a logistic discriminant metric learning method for this challenging problem. Different with most existing metric learning algorithms, it exploits both original data and auxiliary data during training, which is motivated by the new machine learning paradigm - Learning Using Privileged Information. Such privileged information is a kind of auxiliary knowledge which is only available during training. 
Our goal is to learn an optimal distance function by constructing a locally adaptive decision rule with the help of privileged information.
We jointly learn two distance metrics by minimizing the empirical loss penalizing the difference between the distance in the original space and that in the privileged space. In our setting, the distance in the privileged space functions as a local decision threshold, which guides the decision making in the original space like a \textit{teacher}. The metric learned from the original space is used to compute the distance between a probe image and a gallery image during testing. 
In addition, we extend the proposed approach to a multi-view setting which is able to explore the complementation of multiple feature representations. In the multi-view setting, multiple metrics corresponding to different original features are jointly learned, guided by the same privileged information. Besides, an effective iterative optimization scheme is introduced to simultaneously optimize the metrics and the assigned metric weights.
Experiment results on several widely-used datasets demonstrate that the proposed approach is superior to global decision threshold based methods and outperforms most state-of-the-art results.
\end{abstract}

\IEEEpeerreviewmaketitle

\section{Introduction}
\label{Section1}
Person Re-identification (re-ID) \cite{zheng2016Survey} is a critical problem in video analytics applications such as security and surveillance and has attracted increasing attention in recent years.  Although many approaches have been proposed, it remains a challenging problem since person's appearance usually undergoes dramatic changes across camera views due to changes in view angle, body pose, illumination and background clutter.

The fundamental problem is to compare a person of interest from a probe camera view to a gallery of candidates captured from a camera that does not overlap with the probe one. If a `true' match to the probe exists in the gallery, it should have a high matching score, compared to incorrect candidates. Generally speaking, person re-ID involves two sub-problems: feature representation and metric learning. 
An effective feature representation \cite{GaussianReIDDescriptor,LOMOXQDA} is critical for person re-ID, which should be robust to complex variations in human appearances from different camera views. The general strategy is to concatenate multiple low-level visual features into a long feature vector. However, it inevitably brings massive redundant information which may degrade the ability of representation. Therefore, more efforts \cite{ PCCA,MLAPG,MultiTaskDMLReID,zhang2016learning,KISSme,MetricEnsemble,LADFReID,KLFDA,ChenDaPeng_similarityLearning} have been made along the second direction. Some of them formulate re-ID as a subspace learning problem by learning a low-dimensional projection. Some others directly learn a Mahalanobis distance function parameterized by a positive semidefinite (PSD) matrix to separate positive person image pairs from negative pairs. This work follows the second approach, aiming to learn a suitable distance metric.

Despite the promising efforts have been made, most existing metric learning methods are limited in that they compare the distance between a pair of similar/dissimilar instances with a global threshold. Such global threshold based pairwise constraints may suffer from sub-optimal learning performance when coping with some real-world tasks with complex inter-class and intra-class variations, e.g., person re-ID. A natural solution to alleviate this limitation is to design a locally adaptive decision rule. Li et al. \cite{LADFCVPR13} proposed to learn a second-order local decision function in original feature space to replace the global threshold. Wang et al. \cite{AdapMetricECCV14} introduced an adaptive shrinkage-expansion rule to shrink/expand the Euclidean distance as an adaptive threshold. These two earlier works both leverage the information from original feature space to guide the decision making. However, the guidance from the original feature space might be relatively weak, since original feature is usually noisy and less discriminative. It is of great interest to design a solution that is able to exploit additional knowledge beyond the original data space.

It has been shown in a new learning paradigm - Learning Using Privileged Information (LUPI)\cite{JMLR:v16:vapnik15b} that a more reliable and effective model can be learned if some auxiliary expert knowledge is exploited during training. Such auxiliary knowledge is called privileged information and is only available during training. It typically describes some important properties of the training instance, such as attributes, tags, textual descriptions or other high-level knowledge, etc. The LUPI paradigm is inspired by the human teaching-learning in which students will learn better if a teacher can provide some explanations, comments, comparison or other supervision. 

Motivated by the LUPI paradigm, we present a logistic discriminant metric learning method for cross-view re-ID by exploiting both original data and auxiliary data to design a locally adaptive decision rule during training. In our setting, each training instance is represented with two forms of features: one is from the original space and the other is from the privileged space. 
We jointly learn two distance metrics with PSD constraints by minimizing the empirical loss penalizing the difference between the distance in the original space and the distance in the privileged space. 
During training, the distance in the privileged space functions as a local decision threshold to guide the metric learning in the original space like a \textit{teacher}. 
The finally learned metric from the original space is used to compute the distance between a probe image and a gallery image during testing. Moreover, we extend the proposed algorithm from the single-view setting to a multi-view setting which is able to explore the complementation of multiple feature representations. In the multi-view setting, we simultaneously learn multiple distance metrics from different original feature spaces under the guidance of the same privileged knowledge. An effective iterative optimization algorithm is introduced to simultaneously optimize the metrics and the assigned metric weights. 

Our main contributions can be summarized as follows:

(1) We present an effective logistic discriminant metric learning method by exploiting both original data and privileged information to design a locally adaptive decision rule during training. The proposed decision rule is different with the common way in most existing works that compares the distance between a pair of training instances with a global threshold to decide whether they are similar or dissimilar. In this work, such global threshold is replaced with the squared distance in the privileged space. (We term the proposed method as {LDML+}.)

(2) We extend the proposed method to a multi-view setting, which can explore the complementary information of multiple different features effectively. In this work, multiple distance metrics are learned simultaneously from different original feature spaces under the guidance of the same privileged knowledge, in which each metric is learned using a single feature. (We term the proposed multi-view approach as {MVLDML+}.)

(3) We conduct extensive evaluations on several widely-used datasets. Experimental results show that LDML+ is able to improve the performance of global decision threshold based metric learning methods with the help of privileged information and MVLDML+ can outperform most state-of-the-art results.

The proposed LDML+ method was first introduced in previous work \cite{ERMML+}. In comparison with the preliminary version \cite{ERMML+}, we have improvements in the following aspects: 
(1) we regularize the privileged distance metric to control the complexity of model and avoid falling into local optimum; 
(2) we extend the single-view version in previous work to a multi-view setting in this work which can explore multiple original feature representations effectively; 
(3) we present extensive experimental evaluations and analyses to validate the effectiveness of the proposed LDML+ and MVLDML+ methods on several re-ID datasets.

\section{Related Work}
\label{Section2}
\subsection{Metric Learning}
During the past decades, many algorithms have been developed to learn a distance metric. In this subsection, we will briefly review some classical or related distance metric learning works. 

Xing et al. \cite{xing2002distance} proposed to learn a distance metric by minimizing the distance between similar instances while keeping that between dissimilar instances larger than a predefined threshold. Globerson et al. \cite{globerson2005metric} presented a metric learning algorithm by collapsing all examples in the same class to a single point and pushing examples in other classes infinitely far away. Schultz et al. \cite{schultz2004learning} developed a method for learning a distance metric from relative comparison. Davis et al. \cite{ITMLicml-2007} presented an information-theoretic metric learning approach, which formulates the problem as that of minimizing the differential relative entropy between two multivariate Gaussians under pairwise constraints on the distance function. Weinberger et al. \cite{WeinbergerJMLR09LMNN} proposed to learn a Mahalanobis
distance metric for kNN classification with the goal that k-nearest neighbors always belong to the same class while examples from different classes are
separated by a large margin. Guillaumin et al. \cite{LDMLICCV09} designed a logistic discriminant approach to learn a distance metric. Bian et al. \cite{BianTNNLS12} developed a risk minimization framework for metric learning. 
Mignon et al. \cite{PCCA} proposed to learn a distance metrics from sparse pairwise similarity/dissimilarity constraints in high dimensional input space.

Among the above-mentioned algorithms, our work is related to the methods \cite{LDMLICCV09,BianTNNLS12,PCCA} which explore a logistic discriminant approach for metric learning. However, Guillaumin's work \cite{LDMLICCV09} doesn't use any regularization term including the PSD constraint, which easily suffers from overfitting. Bian's work \cite{BianTNNLS12} relies on a strong assumption that the learned metric is bounded, which is too rigid. Besides, these works \cite{LDMLICCV09,BianTNNLS12,PCCA,ITMLicml-2007} all adopt the global threshold based constraints, which easily suffer from sub-optimal learning performance. Compared with them, we learn a PSD metric by exploiting auxiliary information to construct a locally adaptive decision function, which is more robust and shows better performance. 

Zha et al. \cite{ZhaIJCAI09} also presented a metric learning algorithm that exploits auxiliary knowledge during training. However, it's different with our work. In \cite{ZhaIJCAI09}, the authors pre-trained several auxiliary metrics using several auxiliary datasets to assist the metric learning in the source dataset. Different with \cite{ZhaIJCAI09}, we don't exploit any auxiliary datasets during training. In our work, we exploit auxiliary feature representation (privileged information) of training instances during training, and the auxiliary feature representation is not available for testing.

\subsection{Person Re-identification}
Person re-ID aims to retrieve a person of interest across spatially disjoint cameras. It can be seen as a image retrieval problem \cite{zhengTIP14Coupled,zheng2017Survey,zhengTIP2014Lp}. This paper focuses on tackling the person re-ID problem with the proposed metric learning scheme. In this subsection, we will briefly review some related works. Generally speaking, mainstream re-ID works can be roughly categorized into the following groups.

The first group of methods focus on designing discriminative and invariant features for re-ID \cite{LOMOXQDA,SalientColorName,GaussianReIDDescriptor,Invariant_Color_FeaturesReID,VIPeR, SalienceLearning, WhosDescriptors}. 
 Recently, some new proposed feature descriptors have gained good performance, i.e., local maximal occurrence (LOMO) feature \cite{LOMOXQDA}, weighted histograms of overlapping stripes \cite{WhosDescriptors}, and Gaussian of Gaussian (GOG) descriptor \cite{GaussianReIDDescriptor}. The general trend is that the dimensions of feature descriptors are getting higher by concatenating multiple low-level visual features, which may result in the so-called curse of dimensionality. To alleviate this problem, our work provides an effective way to simultaneously explore multiple feature representations. 
 
The second group of methods aim to design discriminative distance functions for recognizing people from disjoint camera views \cite{ChenDaPeng_similarityLearning,KISSme,KLFDA,LFDA,LOMOXQDA,MLAPG,MetricEnsemble,MultiTaskDMLReID,PCCA,zhang2016learning,Yang:2017}. 
 In this group, some works aim to learn a Mahalanobis-like distance metric \cite{MLAPG,PRID_MDLReID,ERMML+}, while some other methods focus on seeking a discriminative subspace \cite{zhang2016learning,KLFDA,RCCA,ZhengWeishi_AsymmetricDistance}. 
These two subgroups actually are closely related. 
We briefly introduce some well-known works as follows. 
Zheng et al. \cite{WeishiZheng_RelativeDistanceReID} formulated re-ID as a relative distance comparison learning problem by maximizing the probability that relevant samples have smaller distance than the irrelevant ones. Liao et al. \cite{MLAPG} proposed a logistic discriminant metric learning approach with PSD constraints and asymmetric sample weight strategy.  Kostinger et al. \cite{KISSme} developed a simple and effective metric learning method by computing the difference between the intra-class and inter-class covariance matrix. As an improvement, Liao et al. \cite{LOMOXQDA} proposed a cross-view quadratic discriminant analysis (XQDA) method by learning a more discriminative distance metric and a low-dimensional subspace simultaneously. Pedagadi et al. \cite{LFDA} applied the local fisher discriminant analysis algorithm to match person images by maximizing the inter-class separability while preserving the multiclass modality, whose kernel version was presented for re-ID in \cite{KLFDA}. In \cite{zhang2016learning}, Zhang et al. proposed to overcome the small-sample-size problem in re-ID by learning a discriminative null space, where the within-class scatter is minimized to zero while maximizing the relative between-class separation simultaneously. 

Different with them, we propose a novel metric learning method for re-ID, which incorporates auxiliary knowledge to guide the metric learning in original feature space. In addition, we also present a multi-view extension which can explore the complementation of multiple feature representations by simultaneously learning multiple metrics. Some existing works \cite{MetricEnsemble,zhao2014learning} have investigated the effects of distance fusion approach for re-ID by a two-stage strategy. They first pretrain several base metrics using different descriptors or different metric learning algorithms, and then combine those base distance functions to obtain the final distance function. Different with them, we present a unified multi-metric learning scheme which can simultaneously learn base metrics and metric weights. 

\subsection{Learning Using Privileged Information}
Our work is motivated by the LUPI \cite{JMLR:v16:vapnik15b} paradigm. We will briefly introduce this learning paradigm in this subsection.

LUPI is a new learning paradigm which was first incorporated into SVM in the form of SVM+ by Vapnik et al. \cite{vapnik2009new}, which uses the
additional (privileged) information as a proxy for predicting the slack variables. It is equivalent to learning an oracle that tells which sample is easy or hard to be predicted. This paradigm has been used for multiple tasks, such as hashing \cite{tianyi2016transfer}, action and event recognition \cite{Niu2016}, information bottleneck learning \cite{Motiian_2016_CVPR}, learning to rank \cite{sharmanska2013learning}, image categorization \cite{li2014exploiting}, object localization \cite{feyereisl2014object}, and active learning \cite{yanimage}, etc.

Recently, two related works \cite{FouadTNNLS13,XuTNNLS15} are proposed to learn a metric using privileged information. 
Fouad et al. \cite{FouadTNNLS13} proposed a two-stage strategy to exploit privileged information for metric learning using the information-theoretic approach\cite{ITMLicml-2007}. They first learn a metric using privileged information to remove some outliers and then use the remaining pairs to learn a metric with original feature. 
Following \cite{FouadTNNLS13}, Xu et al. \cite{XuTNNLS15} proposed the ITML+ method, in which privileged information is used to design a slack function to replace the slack variables in ITML\cite{ITMLicml-2007}. 
Different with these two ITML based methods \cite{FouadTNNLS13,XuTNNLS15}, we provide a new scheme to leverage privileged knowledge for metric learning under a general risk minimization framework.
Moreover, we apply low rank selection for the learned metric in each iteration, which allows us to work directly with higher dimensional input data. 
While ITML based methods aim to learn a full matrix for the target metric that is in the square of the dimensionality, making it computationally unattractive for high dimensional data and prone to overfitting \cite{PCCA}. 
In addition, we present a multi-view extension which is able to simultaneously learn multiple metrics from different original feature spaces, which is also different with \cite{FouadTNNLS13,XuTNNLS15}. 

\section{A Generic Metric Learning Framework}
\label{Section3}
In this section, we introduce a generic metric learning framework which doesn't exploit additional information during training.
Suppose we have a pairwise constrained training set \(\mathcal{Z}=\lbrace\left(\mathbf{x}_i,\mathbf{z}_i,{y}_i\right)|i=1,\cdots,n\rbrace\), where $\mathbf{x}_i\in{\mathbb{R}^d}$, $\mathbf{z}_i\in{\mathbb{R}^d}$ are defined on the same original feature space, $i$ is the index of the $i$-th pair of training instances, and $y_i$ is the label of the pair $\left(\mathbf{x}_i,\mathbf{z}_i\right)$ defined by
\begin{eqnarray}\label{Eq1}
y_{i}=\left \{
\begin{array}{ll}
\;\;\; 1,   \quad  (\mathbf{x}_i,\mathbf{z}_i)\in \mathcal{S}\\
-1, \quad (\mathbf{x}_i,\mathbf{z}_i)\in \mathcal{D} ,
\end{array}
\right.
\end{eqnarray}
where \(\mathcal{S}\) denotes the set of similar pairs and \(\mathcal{D}\) denotes the set of dissimilar pairs. The goal is to learn a Mahalanobis distance metric defined by

\begin{equation}\label{Eq2}
d_\mathbf{M}(\mathbf{x}_i,\mathbf{z}_i)=\sqrt{(\mathbf{x}_i-\mathbf{z}_i)^T\mathbf{M}(\mathbf{x}_i-\mathbf{z}_i)},
\end{equation}
where \(\mathbf{M}\succeq 0\in \mathbb{R}^{d\times d}\) is a PSD distance metric. The learned distance \(d_\mathbf{M}(\mathbf{x}_i,\mathbf{z}_i)\) is expected to be small if \(\mathbf{x}_i\) and  \(\mathbf{z}_i\) are similar, or large if they are dissimilar.

Given a metric, how to determine whether two instances are similar or dissimilar? A common way \cite{MLAPG,LDMLICCV09,hu2014discriminative,PCCA,BianTNNLS12} is to compare their distance with a global decision threshold \(\sigma\). Hence, the decision function $f$ can be defined by
\begin{equation}\label{Eq3}
f(\mathbf{x}_i,\mathbf{z}_i;\mathbf{M})=\sigma-(\mathbf{x}_i-\mathbf{z}_i)^T\mathbf{M}(\mathbf{x}_i-\mathbf{z}_i).
\end{equation}
 If they are similar, the decision function $f>0$, otherwise $f\leq0$. Given the decision function, the problem of metric learning can be cast in a generic framework in which the metric is obtained by minimizing the empirical risk \(J(\mathbf{M})\)
 \begin{equation}\label{Eq4}
\min\limits_{\mathbf{M}\succeq 0} J(\mathbf{M})=\sum\limits_{i=1}^n w_i\mathcal{L}\big (y_{i} f(\mathbf{x}_i,\mathbf{z}_i;\mathbf{M})\big),
\end{equation}
where $\mathcal{L}(\cdot)$ is a loss function that is decreasing monotonically, e.g., log loss and smooth hinge loss. $w_i$ is a weight for the $i$-th pair.
 Existing works \cite{MLAPG,LDMLICCV09,hu2014discriminative,PCCA,BianTNNLS12} are all under this framework.
\section{The proposed approach}
\label{Section4}
\subsection{Problem Formulation}\label{section4.1}
As shown in section \ref{Section3}, traditional pairwise constrained methods only exploit original data during training. They usually adopt the global threshold based decision function, which is too rough to obtain a reasonable metric. In this section, we aim to design a locally adaptive decision rule by exploiting additional knowledge.

Motivated by the LUPI paradigm\cite{JMLR:v16:vapnik15b}, we exploit privileged information to design an adaptive decision function in the training stage. First, each training instance is represented with two forms of features: one is \(\mathbf{x}_i\in{\mathbb{R}^d}\) in the original feature space; the other is \(\mathbf{x}_i^*\in{\mathbb{R}^{d^*}}\) in the privileged space.
The training set is reformulated as \(\mathcal{Z}=\lbrace\left(\mathbf{x}_i,\mathbf{x}_i^*,\mathbf{z}_i,\mathbf{z}_i^*,y_i\right)|i=1,\dots,n\rbrace\). 
 Then, we replace the global threshold $\sigma$ in Eq. (\ref{Eq3}) using the squared distance \(d^2_\mathbf{P}(\mathbf{x}_i^*,\mathbf{z}_i^*)\) in the privileged space, where \(\mathbf{P}\in \mathbb{R}^{{d^*}\times {d^*}}\) is the distance metric corresponding to the privileged information.  Here, \(d_\mathbf{P}^2(\mathbf{x}_i^*,\mathbf{z}_i^*)\) functions as a local decision threshold for the $i$-th training pair. Our locally adaptive decision function is formulated by
\begin{equation}\label{Eq5}
\begin{array}{l}
 \quad f(\mathbf{x}_i,\mathbf{z}_i;\mathbf{x}_i^*,\mathbf{z}_i^*;\mathbf{M},\mathbf{P})\\
=\beta d_\mathbf{P}^2(\mathbf{x}_i^*,\mathbf{z}_i^*)-d_\mathbf{M}^2(\mathbf{x}_i,\mathbf{z}_i)\\
=\beta (\mathbf{x}_i^*-\mathbf{z}_i^*)^T\mathbf{P}(\mathbf{x}_i^*-\mathbf{z}_i^*)-
  (\mathbf{x}_i-\mathbf{z}_i)^T\mathbf{M}(\mathbf{x}_i-\mathbf{z}_i)\\
\end{array},
\end{equation}
where $\beta>0$ is a scale parameter. 
The idea behind Eq. (\ref{Eq5}) is that \textit{teacher}'s concept of similarity between a pair of training instances is usually more credible. We expect the knowledge of \textit{teacher} to be transferred from the privileged space to the original space where decision is made.

With the locally adaptive decision function, our problem can be formulated as
\begin{eqnarray}
\langle\hat{\mathbf{M}},\hat{\mathbf{P}}\rangle=\arg\min J(\mathbf{M},\mathbf{P}) \ \ \ \ \ \ \ s.t.\ \  \mathbf{M}\succeq 0;\mathbf{P}\succeq 0,\label{Eq6}\\
J(\mathbf{M},\mathbf{P})\!=\!\sum\limits_{i=1}^{n} \! w_i \mathcal{L}\big(y_i f \! \left(\mathbf{x}_i,\mathbf{z}_i;\mathbf{x}_i^*,\mathbf{z}_i^*;\mathbf{M},\mathbf{P}\right)\!\big)\!+\!\lambda\mathcal{R}(\mathbf{P}).\label{Eq7}
\end{eqnarray}
As seen from Eq. (\ref{Eq7}), the objective function $J(\mathbf{M},\mathbf{P})$ is constituted by two terms: one is a loss term and the other is a regularization term on $\mathbf{P}$. $\lambda>0$ is a regularization parameter which makes a trade-off between the two terms.

In this work, we denote $\mathcal{L}(\cdot)$ with the log loss
\begin{equation}\label{Eq8}
\mathcal{L}(u)=\ln (1+\exp (-u))
\end{equation}
 which has shown good performance in some existing works. The weight $w_i$ is defined as $\frac{1}{\vert \mathcal{S} \vert}$ if $y_i=1$, or $\frac{1}{\vert \mathcal{D} \vert}$ if $y_i=-1$, in which $\vert \mathcal{S} \vert$ and $\vert \mathcal{D} \vert$ denote the number of similar training pairs and the number of dissimilar training pairs, respectively. The regularization term $\mathcal{R}(\mathbf{P})$ is defined by
\begin{equation}\label{Eq9}
\mathcal{R}(\mathbf{P})=\Vert \mathbf{P} \Vert^2_F/{d^*}.
\end{equation}
where $\Vert \cdot \Vert_F$ denotes the Frobenius norm. Here, the regularization term is added to control the model complexity. It should be noted that we only regularize the metric $\mathbf{P}$, while give the metric $\mathbf{M}$ more freedom. Since once $\mathbf{P}$ has higher degree of freedom than $\mathbf{M}$, the former may be shifted to the latter during training. It should be ensured that $\mathbf{M}$ is guided (supervised) by the privileged information. That is, \textit{student}'s concept of similarity between training instances should be under the control of \textit{teacher}. 

In this subsection, a logistic discriminant metric learning method is presented that exploits auxiliary knowledge to build a locally adaptive decision rule during training. We term this method as \textbf{LDML+} for simplicity.

\subsection{Multi-view Extension}
The proposed LDML+ method only considers a single original features. In this subsection, we extend LDML+ from the single-view setting 
\footnote{Although LDML+ uses two features for training, only the original feature is utilized during testing. Therefore, we categorize it as a single-view method.}
 to a multi-view setting to exploit the complementation of multiple original features.
 
 In the multi-view setting, each training instance is represented by $\mathcal{M}$ original features $\{\mathbf{x}_i^m\in{\mathbb{R}^{d^m}}\}_{m=1}^\mathcal{M}$ and a single privileged feature $\mathbf{x}^*_i\in{\mathbb{R}^{d^*}}$. During training, we aim to simultaneously learn multiple metrics $\mathbf{M}^1,\cdots,\mathbf{M}^\mathcal{M}$ from different original spaces and a single metric $\mathbf{P}$ from the privileged space.
 
Our objective function that is being minimized can be reformulated as 
 \begin{equation}\label{Eq10}
 \begin{aligned}
   & J\left(\mathbf{M}^1,\cdots,\mathbf{M}^\mathcal{M},\mathbf{P},\mathbf{a}\right)\\
 =&\!\sum\limits_{m=1}^{\mathcal{M}}\!\!  a_m^r \Bigg\{\!\sum\limits_{i=1}^{n} \! w_i \mathcal{L}\Big(y_i f \! \left(\mathbf{x}_i^m,\mathbf{z}_i^m;\mathbf{x}_i^*,\mathbf{z}_i^*;\mathbf{M}^m,\mathbf{P}\right)\!\!\Big) \!
+\! \lambda\mathcal{R}(\mathbf{P}) \Biggr\},\\
  & s.t.\ \  \mathbf{M}^m\succeq 0;\mathbf{P}\succeq 0;a_m>0;\sum\limits_{m=1}^{\mathcal{M}} a_m =1
 \end{aligned}
 \end{equation}
where $r>1$ and $\mathbf{a}=(a_1,\cdots,a_m,\cdots,a_\mathcal{M})$ consists of $\mathcal{M}$ weights for $\mathcal{M}$ metrics from original spaces. In Eq. (\ref{Eq10}), we adopt a trick utilized in \cite{wang2009unified,xia2010multiview} that uses $a_m^r$ ($r>1$) instead of $a_m$. It ensures that each view has a particular contribution to the final distance. 

In this subsection, we simultaneously learn multiple view-specific metrics with the help of privileged information.  
We term the proposed multi-view method as \textbf{MVLDML+}. 
\subsection{Solution of Eq. (\ref{Eq10})}\label{section4.3}
In this subsection, we adopt an alternating optimization strategy to solve the minimization problem in Eq. (\ref{Eq10}) with the loss function in Eq. (\ref{Eq8}) and the regularizer in Eq. (\ref{Eq9}). More specifically, we alternatively update $\mathbf{M}^m$($m=1,\cdots,\mathcal{M}$), $\mathbf{P}$, and $\mathbf{a}$ to optimize the objective.
\subsubsection{Optimization of $\mathbf{M}^m$} \label{section3.31}
To optimize $\mathbf{M}^m$, we first fix $\mathbf{P}$, $\mathbf{a}$, and $\mathbf{M}^1$, $\cdots$, $\mathbf{M}^{m-1}$, $\mathbf{M}^{m+1}$, $\cdots$, $\mathbf{M}^{\mathcal{M}}$. Then we derive the derivative of $J$ with respect to $\mathbf{M}^m$ as
\begin{equation}\label{Eq11}
\begin{aligned}
&\frac{\partial}{\partial \mathbf{M}^m} J(\mathbf{M}^1,\cdots,\mathbf{M}^\mathcal{M},\mathbf{P},\mathbf{a})\\
=&\sum\limits_{i=1}^{n} \! \frac{a_m^r w_i y_i \left(\mathbf{x}_i^m-\mathbf{z}_i^m\right)\left(\mathbf{x}_i^m-\mathbf{z}_i^m\right)^T}
{1\!+\!\exp \!\Big(y_i \big(\beta^m d^2_\mathbf{P}\left(\mathbf{x}_i^*,\mathbf{z}_i^*\right)-d^2_{\mathbf{M}^m}\left(\mathbf{x}_i^m,\mathbf{z}_i^m\right)\big)\Big)}
\end{aligned}.
\end{equation}

Based on the derivative $\frac{\partial J}{\partial \mathbf{M}^m}$, we adopt a general gradient descent to update the metric $\mathbf{M}^m$ :
\begin{equation}\label{Eq12}
\mathbf{M}^m_{t}=\mathbf{M}^m_{t-1}-\eta_{t}^m \frac{\partial J}{\partial \mathbf{M}^m}|_{\mathbf{M}^m=\mathbf{M}^m_{t-1}},
\end{equation}
where $\mathbf{M}^m_{t-1}$ denotes the value of $\mathbf{M}^m$ in the ($t-$1)-th iteration. $\eta_{t}^m$ denotes the step-size for $\mathbf{M}^m$ in the $t$-th iteration. In addition, since $\mathbf{M}^m$ is constrained to be PSD ($\mathbf{M}^m\succeq 0$), the output at Eq. (\ref{Eq12}) should further be projected into the PSD cone using \textit{singular value decomposition} (SVD)
\begin{equation}\label{Eq13}
\mathbf{M}^m_{t}=\mathbf{SVD}\bigg(\mathbf{M}^m_{t-1}-\eta_{t}^m \frac{\partial J}{\partial \mathbf{M}^m}|_{\mathbf{M}^m=\mathbf{M}^m_{t-1}}\bigg),
\end{equation}
where only the eigenvectors corresponding to positive eigenvalues are retained in the solution. Therefore, the solution $\mathbf{M}^m_{t}$ has a low-rank structure. It can be factorized into $\mathbf{U} \mathbf{U}^T$ in which $\mathbf{U}$ can be used for dimension reduction.

{During the optimization of $\mathbf{M}^{m}$, we dynamically adapt the step-size $\eta^m$ to accelerate the optimization process, while guaranteeing the convergence. We first initialize the step-size with a large value $\eta^m_0$. At each iteration, we perform the dynamic step-size search strategy to find a suitable step-size $\eta^m$ that satisfies the following condition
\begin{equation}\label{Eq14a}
\begin{aligned}
&J(\mathbf{M}^1,\cdots,\mathbf{M}^{m}_{t},\cdots,\mathbf{M}^\mathcal{M},\mathbf{P},\mathbf{a})\\
<&J(\mathbf{M}^1,\cdots,\mathbf{M}^{m}_{t-1},\cdots,\mathbf{M}^\mathcal{M},\mathbf{P},\mathbf{a})
\end{aligned}.
\end{equation} 
  If the condition Eq. (\ref{Eq14a}) is not satisfied, we shrink the step-size $\eta^m_{t}$ to $\eta^m_{t}/2$ and repeat the operation in Eq. (\ref{Eq13}) until the condition  is satisfied \cite{MLAPG}. 
  When the condition is satisfied, we double the step-size as $\eta^m_{t+1}=2\eta^m_{t}$ for next iteration. The effectiveness of enlarging the step-size in gradient-descent based metric learning has been shown in \cite{JunYu_multiviewDML}. Besides, to avoid the step-size to be enlarged unboundedly, we bound the step-size $\eta^m_t<{s\eta^m_1} (s>1)$ to make the optimization more stable. Here, $\eta^m_1$ is the found step-size which satisfies the condition Eq. (\ref{Eq14a}) at the first iteration. The parameter $s$ controls the upper-boundary of the step-size.}

\subsubsection{Optimization of $\mathbf{P}$} 
Now we consider the optimization of $\mathbf{P}$. Considering $\mathbf{M}^m$($m=1,\cdots,\mathcal{M}$) and $\mathbf{a}$ are fixed, then we derive the derivative of $J$ with respect to $\mathbf{P}$ as
\begin{equation}\label{Eq14}
\begin{aligned}
&\frac{\partial}{\partial \mathbf{P}} J(\mathbf{M}^1,\cdots,\mathbf{M}^\mathcal{M},\mathbf{P},\mathbf{a})\\
=& \!\sum\limits_{m=1}^{\mathcal{M}}\!\!\Bigg\{\!
\sum\limits_{i=1}^{n}  \!
\frac{-a_m^r w_i y_i \beta^m \left(\mathbf{x}_i^*-\mathbf{z}_i^*\right)\left(\mathbf{x}_i^*-\mathbf{z}_i^*\right)^T}
{1\!+\!\exp\!\Big(y_i \big(\beta^m d^2_\mathbf{P}\!\left(\mathbf{x}_i^*,\mathbf{z}_i^*\right)\!-\!d^2_{\mathbf{M}^m}\!\left(\mathbf{x}_i^m,\mathbf{z}_i^m\right)\!\big)\!\Big)}\\
&\ \ \ \ \ \ \ \ +2\frac{\lambda}{d^*} a_m^r \mathbf{P}\Bigg\}.
\end{aligned}
\end{equation}

Based on the derivative $\frac{\partial J}{\partial \mathbf{P}}$, we also adopt the gradient descent and PSD projection to update $\mathbf{P}$ as
\begin{equation}\label{Eq15}
\mathbf{P}_{t}=\mathbf{SVD}\bigg(\mathbf{P}_{t-1}-\eta^*_t \frac{\partial J}{\partial \mathbf{P}}|_{\mathbf{P}=\mathbf{P}_{t-1}}\bigg),
\end{equation}
where $\mathbf{P}_{t-1}$ denotes the value of $\mathbf{P}$ in the ($t-$1)-th iteration. $\eta^*_t $ is the step-size in the $t$-th iteration for $\mathbf{P}$. 
We also apply the dynamic step-size search strategy shown in subsection \ref{section3.31} for the optimization of $\mathbf{P}$. However,
different with that for $\mathbf{M}^{m}$, the step-size $\eta^*_t $ for $\mathbf{P}$ is not allowed to be enlarged during the optimization. 
\subsubsection{Optimization of $\mathbf{a}$} Considering $\mathbf{M}^m$($m=1,\cdots,\mathcal{M}$) and $\mathbf{P}$ are fixed, then the optimization problem at Eq. (\ref{Eq10}) can be transformed as 
\begin{equation}\label{Eq16}
\min \sum\limits_{m=1}^{\mathcal{M}} a^r_m F_m,\quad s.t. \quad a_m>0; \sum\limits_{m=1}^{\mathcal{M}} a_m=1
\end{equation}
where $F_m\!=\!\sum\limits_{i=1}^{n}  w_i \mathcal{L}\Big(y_i f  \left(\mathbf{x}_i^m,\mathbf{z}_i^m;\mathbf{x}_i^*,\mathbf{z}_i^*;\mathbf{M}^m,\mathbf{P}\right)\!\Big)\!+\!\lambda \mathcal{R}(\mathbf{P})$ which can be treated as a constant.

By using a Lagrange multiplier $\xi$ to take $\sum\limits_{m=1}^{\mathcal{M}} a_m=1$ into consideration, we get the objective function as
\begin{equation}\label{Eq17}
Q(\mathbf{a},\xi)={\sum\limits_{m=1}^{\mathcal{M}} a_m^r F_m}-\xi\left(\sum_{m=1}^{\mathcal{M}} a_m-1\right).
\end{equation}

By setting the derivative of $Q(\mathbf{a},\xi)$ with respect to $a_m$($m=1,\cdots,\mathcal{M}$) and $\xi$ to zero
\begin{equation}\label{Eq18}
\left\{\\
\begin{aligned}
\frac{\partial Q}{\partial a_1}&=r a_1^{r-1} F_1-\xi=0\\
\vdots&\\
\frac{\partial Q}{\partial a_\mathcal{M}}&=r a_\mathcal{M}^{r-1} F_\mathcal{M}-\xi=0\\
\frac{\partial Q}{\partial \xi}&=\sum_{m=1}^{\mathcal{M}} a_m-1=0
\end{aligned}\right.,
\end{equation}
we can obtain the closed-form solution of $a_m$($m=1,\cdots,\mathcal{M}$)
\begin{equation}\label{Eq19}
a_m=\frac{\left(1/F_m\right)^{\frac{1}{r-1}}}{\sum\limits_{m=1}^{\mathcal{M}}\left(1/F_m\right)^{\frac{1}{r-1}}},
\end{equation}
Since $F_m$ is positive, we have $a_m>0$ naturally. When $\mathbf{M}^m$($m=1,\cdots,\mathcal{M}$) and $\mathbf{P}$ are fixed, Eq. (\ref{Eq19}) gives the global optimal $\mathbf{a}$.

According to Eq. (\ref{Eq19}), we have the following understanding for $r$ in controlling $a_m$.  If $r\rightarrow \infty$, $a_m$ will close to each other, i.e., $a_m\rightarrow \frac{1}{\mathcal{M}}$ and each view has an equal contribution. If $r\rightarrow 1$, only $a_m$ corresponding to the minimum $J_m$ is close to 1; other weights will be close to zero. Therefore, the choice of $r$ should be based on the complementary property of all views. Rich complementary prefers to a large $r$; otherwise, it should be small.

Note that we doesn't provide a separate solution for LDML+, since it can be seen as a special case of MVLDML+. 
\begin{figure}[tbp]
	\begin{center}
		\includegraphics[width=3.3in]{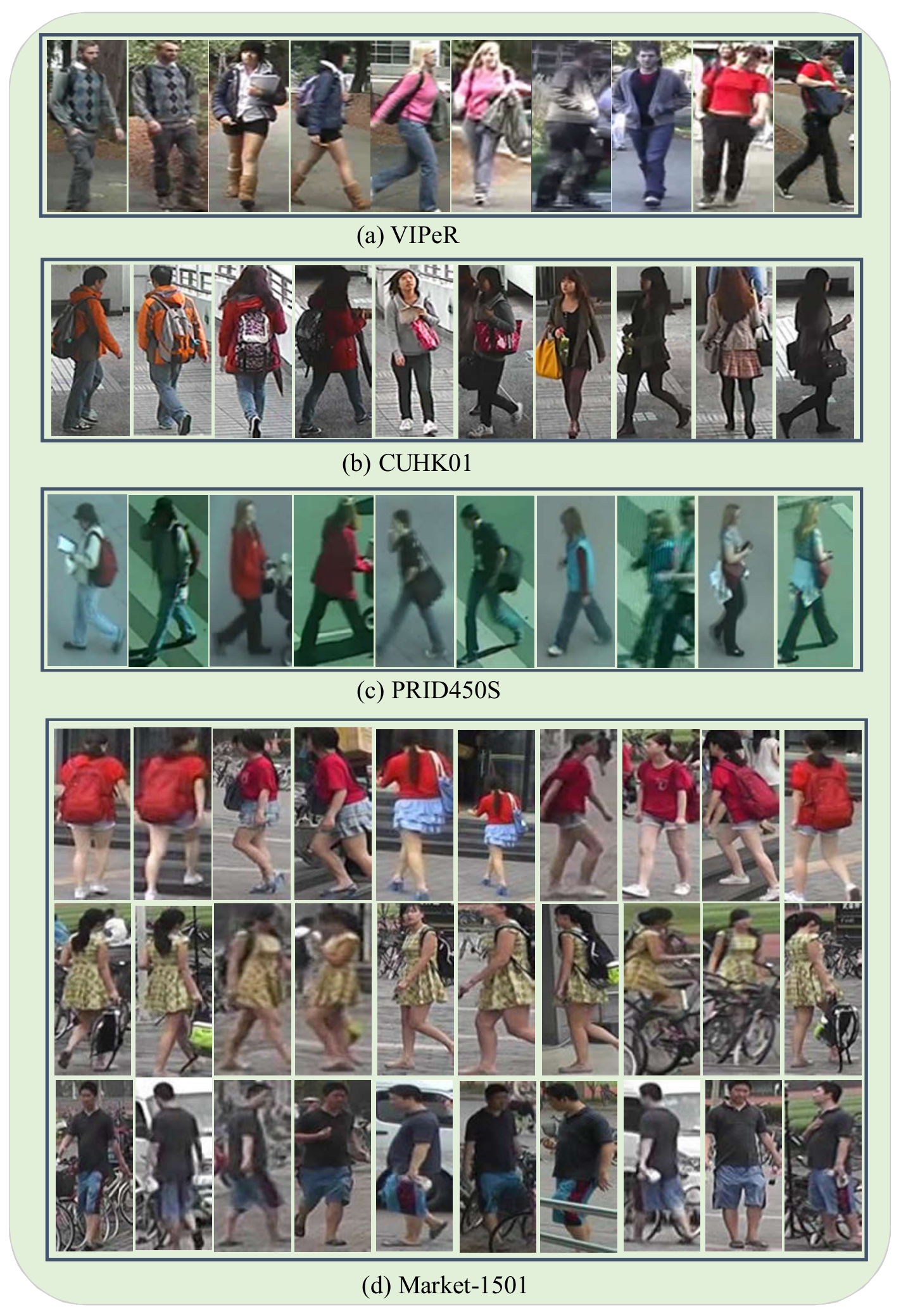}
	\end{center}
	\vspace{-0.1500in}
	\caption{Sample images from four person re-identification datasets: (a) VIPeR; (b) CUHK01; (c) PRID450S; (d) Market-1501.}
	\vspace{-0.050in}
	\label{fig1}
\end{figure}	
\subsection{Person Re-ID}
Once metrics $\mathbf{M}^m$($m=1,\cdots,\mathcal{M}$) and weights $\mathbf{a}$ have been learned after optimization, given a probe image $\mathbf{x}^p$ and a set of gallery images $\{\mathbf{x}^g_j\}_{j=1}^N$ during testing, we compute their squared distances as follows to perform person images matching.
\begin{equation}\label{Eq.21}
\begin{aligned}
d^2\left(\mathbf{x}^p,\mathbf{x}^g_j\right)
&=\sum\limits_{m=1}^{\mathcal{M}} a_m{(\mathbf{x}^p-\mathbf{x}^g_j)^T \mathbf{M}^m (\mathbf{x}^p-\mathbf{x}^g_j)}\\
&=\sum\limits_{m=1}^{\mathcal{M}} a_m{\Vert {(\mathbf{U}^m)}^T\mathbf{x}^p-{(\mathbf{U}^m)}^T\mathbf{x}^g_j\Vert^2}
\end{aligned},
\end{equation}
where $\mathbf{U}^m$ is the low-dimensional projection of $\mathbf{M}^m$, obtained by SVD in Eq. (\ref{Eq13}).
The gallery images can be ranked according to their distances to the probe image. 

\begin{table}[tbp]
	\caption{The characteristics of four person re-ID datasets. }
		\renewcommand\arraystretch{1.2}
		\begin{center}
		\begin{tabular}{l|c|c|c|c}
	\ChangeRT{0.7pt}
	{Datasets}      & {\# ID}& {\# BBoxes} & {\# Distra} & {\# Cam}  \\
	\hline\hline
	VIPeR \cite{VIPeR}         &632 &1264 &0 &2 \\
	CUHK01\cite{li2012human}    &971 & 3884&0 & 2\\
	PRID450S\cite{PRID_MDLReID} &450 & 900& 0&2 \\
	Market-1501\cite{ZhengLiangReIDBenchmark} &1501 & 32668& 2793&6\\
	\ChangeRT{0.7pt}		  			
\end{tabular}
		\end{center}
	\label{Table1}
\end{table}	
\section{Experimental Results and Analysis}
\subsection{Datasets, Evaluation Protocol, and Setting}\label{5.1}
\subsubsection{Datasets}
The evaluation of the proposed methods is carried out on four benchmark person re-ID datasets: VIPeR \cite{VIPeR}, CUHK01\cite{li2012human},  PRID450S\cite{PRID_MDLReID}, and {Market-1501 \cite{ZhengLiangReIDBenchmark}}.

VIPeR \cite{VIPeR} is the most commonly used person re-ID dataset containing 632 persons in which each person has a pair of images taken from widely differing views. It contains 1264 images in total. The large viewpoint change of 90 degrees or more as well as huge lighting variations in VIPeR make it one of the most challenging re-ID datasets. The evaluation protocol for VIPeR is to randomly split the dataset into half, 316 persons for training and 316 persons for testing. 

CUHK01 \cite{li2012human} contains 971 persons from two disjoint camera views, where each person has two images in each camera view. It contains 3884 images in total. We randomly partition the CUHK01 dataset into 486 persons for training and 485 for testing. 

PRID450S\cite{PRID_MDLReID} is a commonly used dataset which contains 450 identities from two disjoint camera views, where each person has one image in each camera view. 225 persons are randomly selected for training and the rest for testing.

{Market-1501} \cite{ZhengLiangReIDBenchmark} is one of the largest person re-ID datasets, containing 32668 bounding boxes (cropped images) of 1501 identities. All the bounding boxes are detected by Deformable Part Model pedestrian detector \cite{cho2012real}.
	Each identity has multiple images captured by at least two cameras and at most six cameras. We adopt the standard protocol \cite{ZhengLiangReIDBenchmark} for Market-1501. Specifically, the training set contains 12936 bounding boxes of 750 identities. The testing set contains 19732 bounding boxes of 751 identities, where only one image of each identity is randomly selected as query image for each camera. In total, the testing set contains 3368 query images. There are 2793 images included as distractors in the original gallery set for testing.

Table \ref{Table1} provides a statistical summary of each dataset. In Table \ref{Table1}, we indicate the number of identities (ID), bounding boxes (BBoxes), distractors (Distra), and cameras (Cam) in each dataset. Fig. \ref{fig1} shows some image samples from these four datasets.
\subsubsection{Evaluation protocol}
For the three small datasets: VIPeR, CUHK01, and PRID450S, we randomly divide each dataset into training and testing sets containing half of the available individuals. As random selection is involved, we repeat the evaluation procedure for 10 times and report the mean results. Single-query (SQ) setting is adopted for all these three datasets. 
For the Market-1501 dataset, one of the largest re-ID datasets, we use its standard evaluation protocol \cite{ZhengLiangReIDBenchmark}.
Both the single-query and multi-query (SQ) matching results are reported on Market-1501. 
Two standard evaluation metrics are used in this work: Cumulative Matching Characteristics (CMC) curve and Mean Average Precision (mAP). 
The CMC curve provides a ranking for every image in the gallery with respect to the probe. It is used for all datasets. The mAP measure is only presented for Market-1501. 

\subsubsection{Features} {The GOG descriptor \cite{GaussianReIDDescriptor}  describes a local region in a person image via hierarchical Gaussian distribution in which both means and covariances are included in their parameters. It has four GOG features: \{GOG$_\textrm{RGB}$, GOG$_\textrm{Lab}$, GOG$_\textrm{HSV}$, GOG$_\textrm{nRnG}$\}, extracted respectively from four color channels \{RGB, Lab, HSV, nRnG\}. Here, the nRnG is the normalized RGB color space.
In this work, we evaluate the effectiveness of LDML+ using the first three GOG features as original features respectively. All the GOG features are used to evaluate MVLDML+.

For the privileged information, it is better to be represented by some high-level features (e.g., attributes). However, up to now, it is hard to obtain ideal privileged features. 
To evaluate the effectiveness of the proposed method in this work, we consider an approximated setting for privileged information. For the VIPeR, CUHK01, and PRID450S datasets, we fuse multiple strong visual features (LOMO feature and FTCNN feature\cite{matsukawa2016person}) to approximate the privileged information. We apply the method in \cite{LOMOXQDA} to obtain a low-dimensional representation of the approximated privileged feature. For the Market-1501 dataset, we combine the predicted pedestrian attributes  \cite{lin2017improving} and semantics-preserving deep embeddings \cite{lin2017improving} as the privileged information. 
Note that the privileged information is not available during testing.}

\subsubsection{Setting}
 For the scale parameter in Eq. (\ref{Eq5}), we set $\beta=\frac{mean(\mathbf{D}_\mathbf{M})
 }{mean(\mathbf{D}_\mathbf{P})}$ where $\mathbf{D}_\mathbf{M}, \mathbf{D}_\mathbf{P}$ denote the squared Euclidean distance matrices corresponding to the original features and the privileged features, respectively. The regularization parameter $\lambda$ is empirically set to 0.0001 for Market-1501. For other datasets, $\lambda$ is set to 0.001.
We set the parameter $r=3$ in Eq. (\ref{Eq10}) empirically. For the optimization of $\mathbf{M}$ and $\mathbf{P}$, we initialize the step-sizes with large values as $\eta^m_0=2^{20}$ and $\eta'_0=2^{15}$ respectively. For the parameter
 $s$ which controls the upper-boundary of $\eta^m$, we set it to $s=2^5$. The maximal iteration number is set to 400 with a stopping criterion by $\vert \frac{J_{t}-J_{t-1}}{J_{t-1}}\vert  \leq 10^{-4}$. { PCA is applied for all datasets for dimension reduction but all energies are retained for the three small datasets. After PCA, the detailed dimensions of the original features on VIPeR, CUHK01, and PRID450S are 631, 1943, and 449 respectively. For the large dataset, Market-1501, we retain 99\% energies. After PCA on Market-1501, the detailed dimensions of the GOG$_\textrm{RGB}$, GOG$_\textrm{Lab}$, GOG$_\textrm{HSV}$ features are 4411, 4409, and 4468, respectively.}
 
 \begin{table*}[htbp]
  						\caption{Top-ranked average recognition rates (CMC@rank-r, \%) of LDML+ and four baseline methods on the \textbf{VIPeR}, \textbf{CUHK01}, and \textbf{PRID450S} datasets with three different visual feature descriptors. A larger number indicates a better result. The best results are shown in boldface.}
  							\begin{center}	
   						\renewcommand\arraystretch{1.2}
  			\begin{tabular}{c|l  |c c c c  ||c c c c ||c c c c} 
  						\ChangeRT{0.7pt}
  				 \multicolumn{2}{c|} {\multirow{2}{*}{VIPeR } } &\multicolumn{4}{c||}{GOG$_\textrm{RGB}$} &\multicolumn{4}{c||}{GOG$_\textrm{Lab}$}  &\multicolumn{4}{c}{GOG$_\textrm{HSV}$}\\\cline{3-14}
  				   		    		    		                   \multicolumn{2}{c|}{}& R=1 & R=5 & R=10 & R=20 & R=1 & R=5 & R=10 & R=20 & R=1 & R=5 & R=10 & R=20\\
  				 					    \hline
  				           	&  XQDA\cite{LOMOXQDA} & 43.77 & 74.81 & 84.34 & 93.32& 44.24& 74.91 & 85.13& 92.94& 39.30& 68.39& 79.59& 89.68\\
  				 				Baseline&  MLAPG\cite{MLAPG} & 42.66& 74.40& 85.47& 93.83& 44.30& 75.38& 85.66& 93.51& 39.59&69.08& 80.82& 89.94 \\
  				 			methods&	LDML$^{\sigma=1}$   & 43.26& 74.84& 85.54& 93.86& 44.08& 75.51& 85.82& \textbf{93.67}&39.46& 69.56 & 80.57& 90.19\\
  				 			&	LDML   & 43.48 & 75.09 & 85.63& 93.77& 44.49 & 75.95& 86.08 & 93.64& 39.46& \textbf{69.68}& \textbf{80.89}& \textbf{90.51}\\
  				 				\hline
  				 			Ours&	LDML+ & \textbf{45.19} & \textbf{75.32} & \textbf{85.66}& \textbf{93.99} & \textbf{45.79} & \textbf{76.33} & \textbf{86.30}& {93.16} & \textbf{40.00} & {69.40} & {80.79} & {89.94} \\
  				 \hline\hline
  				 			  		  \multicolumn{2}{c|} {\multirow{2}{*}{CUHK01}} &\multicolumn{4}{c||}{ GOG$_\textrm{RGB}$} &\multicolumn{4}{c||}{ GOG$_\textrm{Lab}$}  &\multicolumn{4}{c}{GOG$_\textrm{HSV}$}\\\cline{3-14}
  				 			\multicolumn{2}{c|}{}& R=1 & R=5 & R=10 & R=20 & R=1 & R=5 & R=10 & R=20 & R=1 & R=5 & R=10 & R=20\\
  				 			\hline
  				 			&  XQDA\cite{LOMOXQDA} & 55.84 & 78.85 & 85.52 & 91.32 & 53.82 & 77.73& 84.76 & 90.92& 45.55 & 71.46& 79.96 & 87.65\\
  				 			Baseline  & MLAPG\cite{MLAPG}     & 53.38 & 77.72 & 85.19 & 91.29 & 52.77 & 77.32& 85.06& 91.32& 44.72& 71.49& 80.89 & 88.47 \\
  				 			methods &	LDML$^{\sigma=1}$   & 50.01 & 74.89 & 83.58 & 89.94   & 48.31& 72.65 & 81.58 & 89.07 & 39.96 & 65.87 & 75.95 & 85.38 \\
  				 			&	LDML & 54.61 & 78.49 & 85.88 & 91.53  & 53.34 & 77.40& 85.20 & 91.42& 45.47& 71.51& 81.16& 88.67\\
  				 			\hline
  				 			Ours &	LDML+ & \textbf{58.18} & \textbf{80.97} & \textbf{87.45}& \textbf{92.87}  & \textbf{56.88} & \textbf{80.25} & \textbf{87.08}& \textbf{92.47} & \textbf{49.56} & \textbf{75.35} & \textbf{83.60} & \textbf{90.94} \\
  				 		\hline\hline
  				 			   		    \multicolumn{2}{c|} {\multirow{2}{*}{PRID450S }}    &\multicolumn{4}{c||}{ GOG$_\textrm{RGB}$} &\multicolumn{4}{c||}{ GOG$_\textrm{Lab}$}  &\multicolumn{4}{c}{GOG$_\textrm{HSV}$}\\\cline{3-14}
  				 			\multicolumn{2}{c|}{}& R=1 & R=5 & R=10 & R=20 & R=1 & R=5 & R=10 & R=20 & R=1 & R=5 & R=10 & R=20\\
  				 			\hline
  				 			&XQDA\cite{LOMOXQDA} &\textbf{62.36} & \textbf{85.47}& \textbf{91.78} & \textbf{96.27} & \textbf{62.40} & \textbf{86.22} & {92.53} & \textbf{96.98}& {56.00} & {80.93} & {89.02} & {94.71} \\
  				 			Baseline&	  MLAPG\cite{MLAPG}     & 59.56 & 83.47& 90.89 & 95.96 & 58.71& 84.22& 92.13 & 96.40& 53.82 & 79.69& 88.09 & 94.31 \\
  				 			methods&	LDML$^{\sigma=1}$   & 58.62& 83.11& 91.20 & 95.60 & 55.96 & 81.29 & 90.94 & 95.51& 48.62& 75.42 & 85.07& 91.64\\
  				 			&LDML & 59.73 & 83.64& 91.33& 96.04 & 59.64& 84.49 & 92.40 & 96.53 & 54.58& 80.27& 88.44& 94.44\\
  				 			\hline
  				 			Ours&	LDML+ & 60.71 & 84.58  & 91.73& {96.18}  & 60.00& 85.38 & \textbf{92.58}& 96.58  & \textbf{56.49} & \textbf{81.73}  & \textbf{89.42}  & \textbf{95.24}  \\
  				 			\ChangeRT{0.7pt}		
  	\end{tabular}
  		\end{center}
  		\label{Table2}
  		 \vspace{-0.10in}
  		\end{table*}	

\begin{table*}[tbp]
	\caption{Top-ranked average recognition rates (CMC@rank-r, \%) of MVLDML+ and two baseline methods on the \textbf{VIPeR}, \textbf{CUHK01}, and \textbf{PRID450S} datasets. A larger number indicates a better result. The best results are shown in boldface.}
	\begin{center}	
		\renewcommand\arraystretch{1.2}
		\begin{tabular}{l |p{1.85cm} |p{0.56cm} p{0.56cm} p{0.56cm} p{0.56cm} ||p{0.56cm} p{0.56cm} p{0.56cm} p{0.56cm} ||p{0.56cm} p{0.56cm} p{0.56cm} p{0.56cm} } 
			\ChangeRT{0.7pt}
			\multirow{2}{*}{GOG Features }&\multirow{2}{*}{Methods }& \multicolumn{4}{c||}{ VIPeR}  &\multicolumn{4}{c||}{CUHK01} &\multicolumn{4}{c}{PRID450S} \\
			\cline{3-14}
			& & \textrm{  R=1} & \textrm{ R=5} &\textrm{R=10} & R=20 & \textrm{  R=1} & \textrm{ R=5} &\textrm{R=10} & R=20 & \textrm{  R=1} & \textrm{ R=5} &\textrm{R=10} & R=20 \\  				 		 	
			\hline	
			\multirow{3}{*}{\textrm{RGB}, \textrm{Lab}, \textrm{HSV}} 
			& Ensem-XQDA& 47.72& 76.93& 86.99 & 94.75   & 57.59& 79.86 & 86.47 & 92.42    & 64.53& 87.24& 93.02& 96.98\\
			& Ensem-MLAPG & 47.47& 78.01 & 87.59 & \textbf{94.97} & 56.95& 80.33 & 87.11 & {92.77} & 62.76& 85.60 & 92.40& {96.62} \\  \cline{2-14}
			&  {MVLDML+} & \textbf{48.86} & \textbf{78.16} & \textbf{87.82}& {94.53} & \textbf{60.73} & \textbf{82.66} & \textbf{88.99}& \textbf{93.84}  & \textbf{64.80} & \textbf{88.13} & \textbf{94.00}& \textbf{97.64}  \\ 
			\hline
			\multirow{3}{*}{\textrm{RGB}, \textrm{Lab}, \textrm{HSV}, \textrm{nRnG}}  
			&Ensem-XQDA& 49.08 & 77.47& 87.37& 94.62 & 58.43 & 80.04& 86.70& 92.49& \textbf{68.00}& 88.00& 94.00& 97.29\\
			& Ensem-MLAPG & 49.05& 78.67& 88.48 & \textbf{94.97} & 57.75& 80.21& 87.15& {92.60} & 64.36& 86.93 & 93.51& {97.29}\\  \cline{2-14}
			&  {MVLDML+} & \textbf{50.03} & \textbf{79.21} & \textbf{88.54}& {94.65}  & \textbf{61.37} & \textbf{82.74} & \textbf{88.88}& \textbf{93.85} & {66.71} & \textbf{88.80} & \textbf{94.44}& \textbf{97.51}  \\ 	
			\ChangeRT{0.7pt}
		\end{tabular}
	\end{center}
	\label{Table3}
		 \vspace{-0.10in}
\end{table*}
For the evaluation of LDML+, the following baseline methods are employed:
\begin{itemize}
\item [1)] XQDA \cite{LOMOXQDA}.
It is an efficient yet effective metric learning algorithm which learns a discriminative distance metric and a low-dimensional subspace simultaneously. It's a state-of-the-art method especially on small-size datasets.
Default setting in \cite{LOMOXQDA} is applied for XQDA. Note that it doesn't apply PCA for dimension reduction. The input feature vectors will be projected into a low-dimensional subspace directly.
\item [2)] Global decision threshold based methods:
\begin{itemize}
\item MLAPG\cite{MLAPG}. It is a state-of-the-art logistic discriminant metric learning method under Eq. (\ref{Eq4}) in which the global decision threshold $\sigma$ is set to the average squared Euclidean distance. Different with LDML+, it applies the Accelerated Proximal Gradient (APG) algorithm to optimize the metric. We use the default setting in \cite{MLAPG} for MLAPG. 
\item LDML. A logistic discriminant metric learning method under Eq. (\ref{Eq4}) in which the default setting of $\sigma$ is the average squared Euclidean distance. LDML shares the same settings with LDML+ on the loss function $\mathcal{L}(\cdot)$, pair weights $w$, and optimization strategy of $\mathbf{M}$. We also provide the results of LDML with $\sigma=1$ and term it as LDML$^{\sigma=1}$.
\end{itemize}
\end{itemize}	

For the evaluation of MVLDML+, we build two baseline methods using XQDA\cite{LOMOXQDA} and MLAPG\cite{MLAPG}, respectively, based on a score-level fusion strategy. In detail, we learn an ensemble of distance functions, in which each base distance function is learned using a single feature descriptor. The final distance is calculated from a weighted sum of these distance functions with equal weights. The two baseline methods are termed as:
\begin{itemize}
\item  Ensem-XQDA
\item  Ensem-MLAPG
\end{itemize}	
\subsection{Experiments on VIPeR, CUHK01, and PRID450S}
In this subsection, we evaluate the effectiveness of LDML+ and MVLDML+ on the three small-size datasets: VIPeR, CUHK01, and PRID450S, respectively. 
 
\subsubsection{Evaluation of LDML+} 
We first evaluation the effectiveness of LDML+ using three GOG descriptors (GOG$_\textrm{RGB}$, GOG$_\textrm{Lab}$, and GOG$_\textrm{HSV}$), respectively. The evaluation results are shown in Table \ref{Table2}. 

It can be seen from Table \ref{Table2} that the proposed LDML+ method performs better than the three global decision threshold based methods (MLAPG,  LDML$^{\sigma=1}$, and LDML).  Among them, LDML performs better than MLAPG and LDML$^{\sigma=1}$.
On VIPeR, LDML+ surpasses its counterpart, LDML, by 1.71\%, 1.3\%, and 0.54\% at rank-1 using three different GOG features, respectively. On CUHK01, LDML+ beats LDML by a large margin, 3.57\%, 3.54\%, 4.09\% at rank-1, respectively. On PRID450S, the improvements are similar with those on VIPeR, which are 0.98\%, 0.36\%, and 1.91\% at rank-1.
It reveals that the locally adaptive decision rule can cope better with the complex intra-class and inter-class variations than the global threshold based decision rule, especially on the CUHK01 dataset that is larger than the VIPeR and PRID450S datasets. 

Our LDML+ method also outperforms the XQDA method on VIPeR and CUHK01 datasets, benefiting from the locally adaptive decision rule. On VIPeR, the improvements are 1.42\%, 1.55\%, and 0.7\% at rank-1, respectively. On CUHK01, LDML+ yields significant improvements on XQDA at rank-1 by 2.34\%, 3.06\%, and 4.01\%, respectively. XQDA performs better than our method using the GOG$_\textrm{RGB}$ and GOG$_\textrm{Lab}$ features on PRID450S. Since XQDA learns a discriminative low-dimensional subspace and a Mahalanobis distance metric simultaneously. Here, the discriminative subspace projection is a supervised technique for dimensionality reduction, which can retain more discriminative information than the PCA technique used in our method and other baselines. The advantage of a supervised dimensionality reduction technique will be more obvious when it is applied on a small dataset, e.g., PRID450S.

We note that LDML performs better than MLAPG on the three datasets although they both use the mean squared Euclidean distance as the global threshold, which can be owned to the effectiveness of the dynamical step-size adaption strategy applied in LDML+, LDML$^{\sigma=1}$, and LDML. MLAPG also applies a linear search strategy to find a suitable step-size. However, different with ours, it doesn't allow the step-size to be enlarged. Once a very small step-size is searched at the beginning of the optimization, the subsequent gradient descent would be much slower, thus resulting in suboptimal performance.
In addition, as shown in Table \ref{Table2}, compared with LDML$^{\sigma=1}$, LDML achieves better or comparable performances. It demonstrates that a data-dependent global threshold is better than a data-independent global threshold.

From the results shown in Table \ref{Table2}, we can conclude that a more reliable metric can be learned by exploiting the privileged information to design a locally adaptive decision rule during training.

   	\begin{table}[tbp]
   	  						\caption{Comparisons of top-ranked average recognition rates (CMC@rank-r, \%) of MVLDML+ with state-of-the-art results on the \textbf{VIPeR} dataset. A larger number indicates a better result. The best results are shown in boldface.}
   	  							\begin{center}	
   	   						\renewcommand\arraystretch{1.2}
   	  			\begin{tabular}{l|c|c c c c } 
   	 		\ChangeRT{0.7pt}
   	  				 Methods  & References & \textrm{R=1} & \textrm{R=5} &\textrm{R=10} & R=20 \\  				 		 	
   	  		\hline	
   	  				 				{MVLDML+} & Ours & \textbf{50.0} & {79.2} & {88.5}& {94.7}  \\
   	  				 				GOG$_\textrm{Fusion}$+XQDA \cite{GaussianReIDDescriptor}& CVPR 2016 & {49.7} & \textbf{79.7} & {88.7}& {94.5}  \\
   	  				 			      Cheng et al.\cite{cheng2016person} & CVPR 2016 & {47.8} & {74.7} & {84.8}& {91.1}  \\
   	  				 			      GOG$_\textrm{Fusion}$+LDNS\cite{zhang2016learning} & CVPR 2016 & 47.6& 78.1& 88.4& 94.6\\
   	  				 			      MetricEnsemble\cite{MetricEnsemble} & CVPR 2015 & {45.9} & {77.5} & \textbf{88.9}& \textbf{95.8}  \\ 
   	  				 			      mFilter+LADF\cite{zhao2014learning} & CVPR 2014 & 43.4 & 73.0 & 84.9 & 93.7\\
   	  				 			      MirrorKMFA\cite{chen2015mirror} & IJCAI 2015 & 43.0 & 75.8 & 87.3 & 94.8\\
   	  				 			      LSSCDL \cite{HuchuanLu_Sample-Specific_SVM_Learning} & CVPR 2016 & 42.7 & {-} & 84.3 & 91.9\\
   	  				 			      LOMO+LDNS\cite{zhang2016learning} & CVPR 2016 & 42.3 & 71.5 & 82.9 & 92.1\\
   	  				 			      Su et al. \cite{su2015multi} & ICCV 2015 & 42.3 & 72.2 & 81.6 & 89.6\\
   	  				 			      Shi et al.\cite{shi2015transferring} & CVPR 2015 & 41.6 & 71.9 & 86.2 & 95.1\\  				 			 
   	   				 			  	 LOMO+MLAPG\cite{MLAPG}  & ICCV 2015 & 40.7& {-} & 82.3 & 92.4 \\  
   	   				 			  	 LOMO+XQDA\cite{LOMOXQDA} & CVPR 2015 & 40.0 & 68.0 & 80.5 & 91.1 \\
   	   				 			  	 Xiao et al. \cite{xiao2016learning} & CVPR 2016 & 38.6 &{-} &{-} &{-}\\
   	   				 			  	 Chen et al.\cite{DeepRankreID-TIP2016} & TIP 2016 & 38.4 & 69.2 & 81.3 & 90.4 \\
   	   				 			  	 SCNCD\cite{SalientColorName} & ECCV 2014 & 37.8 & 68.5 & 81.2 & 90.4\\
   	   				 			  	 Chen et al. \cite{chen2015similarity} & CVPR 2015 & 36.8 & 70.4 & 83.7 & 91.7\\
   	   				 			  	 Shen et al.\cite{shen2015person} & ICCV 2015 & 34.8 & 68.7 & 82.3 & 91.8\\
   	 			\ChangeRT{0.7pt}		  			
   	  	\end{tabular}
   	  		\end{center}
   	  		\label{Table4}
   	  		\end{table}	

	\begin{table}[tbp]
	\caption{Comparisons of top-ranked average recognition rates (CMC@rank-r, \%) of MVLDML+ with state-of-the-art results on the \textbf{CUHK01} dataset. A larger number indicates a better result. The best results are shown in boldface.}
	\begin{center}	
		\renewcommand\arraystretch{1.2}
		\begin{tabular}{l|c|c c c c} 
			\ChangeRT{0.7pt}		
			{Methods} & {References} & \textrm{R=1} & \textrm{R=5} &\textrm{R=10} & R=20 \\  				 		 	
			\hline	
			{MVLDML+} & Ours                                  & \textbf{61.4} & {82.7} & {88.9}& {93.9}     \\
			GOG$_\textrm{Fusion}$+LDNS\cite{zhang2016learning}        & CVPR 2016 & {60.8}& {81.7} & {88.4} & {93.5}           \\ 
			CPDL  \cite{ReID_projectiveDictionary_IJCAI}    & IJCAI 2015 & {59.5} & {81.3} & {89.7} & {93.1}    \\ 	  
			GOG$_\textrm{Fusion}$+XQDA \cite{GaussianReIDDescriptor}  & CVPR 2016 & {57.8}& {79.1} & {86.2} & {92.1}                     \\ 
			Cheng et al.\cite{cheng2016person}     & CVPR 2016 & {53.7}  & \textbf{84.3} & \textbf{91.0} & \textbf{96.3}   \\  	
			MetricEnsemble\cite{MetricEnsemble} & CVPR 2015 & {53.4} & {76.4} & {84.4} & {90.5}    \\ 
			Chen et al.\cite{DeepRankreID-TIP2016} & TIP 2016 & 50.4 & 75.9 & 84.0 & 91.3  \\ 			 			
			LOMO+XQDA\cite{LOMOXQDA}              & CVPR 2015  & {49.2} & {75.7} & {84.2} & {90.8}   \\
			Ahmed et al. \cite{ahmed2015improved} & CVPR 2015  & {47.5} & {71.0} & {80.0} & {-}   \\
			MirrorKMFA\cite{chen2015mirror}         & IJCAI 2015 & 40.4 & 64.6 & 75.3 & 84.1  \\ 		  				 			
			\ChangeRT{0.7pt}		  			
		\end{tabular}
	\end{center}
	\label{Table5}
\end{table}

\begin{table}[tbp]
	\caption{Comparisons of top-ranked average recognition rates (CMC@rank-r, \%) of MVLDML+ with state-of-the-art results on the \textbf{PRID450S} dataset. A larger number indicates a better result. The best results are shown in boldface.}
	\begin{center}	
		\renewcommand\arraystretch{1.2}
		\begin{tabular}{l|c|c c c c } 
			\ChangeRT{0.7pt}	
			Methods  & References & \textrm{R=1} & \textrm{R=5} &\textrm{R=10} & R=20 \\  				 		 	
			\hline
			{MVLDML+} & Ours & {66.8} & \textbf{88.8} & \textbf{94.8}& {97.7}  \\
			GOG$_\textrm{Fusion}$+XQDA \cite{GaussianReIDDescriptor}  & CVPR 2016 & \textbf{68.4}& \textbf{88.8} & {94.5} & \textbf{97.8}                     \\ 
			FFN\cite{wu2016enhanced} & WACV 2016& {66.6} & {86.8} & {92.8}& {96.9}\\
			GOG$_\textrm{Fusion}$+LDNS\cite{zhang2016learning}        & CVPR 2016 & {64.8}& {88.1} & {94.0} & {97.6}           \\ 
			LOMO+XQDA\cite{LOMOXQDA} & CVPR 2015 & 62.6 & 85.6 & 92.0 & 96.6 \\	
			LSSCDL \cite{HuchuanLu_Sample-Specific_SVM_Learning} & CVPR 2016 & 60.5 & {-} & 88.6& 93.6\\
			MirrorKMFA\cite{chen2015mirror} & IJCAI 2015 & 55.4 & 79.3 & 87.8 & 93.9\\
			MEDVL\cite{Metric_EmbeddedDictonaryReID} & AAAI 2016 & 45.9 & 73.0 & 82.9 & 91.1\\
			Shi et al.\cite{shi2015transferring} & CVPR 2015 & 44.9 & 71.7 & 77.5 & 86.7\\  
			Shen et al.\cite{shen2015person} & ICCV 2015 & 44.4 & 71.6 & 82.2 & 89.8\\		
			SCNCD\cite{SalientColorName} & ECCV 2014 & 41.6 & 68.9 & 79.4 & 87.8\\	  				 			    		
			\ChangeRT{0.7pt}		  			
		\end{tabular}
	\end{center}
	\label{Table6}
\end{table}

\subsubsection{Evaluation of MVLDML+} 
By extending the LDML+ method to the multi-view setting in Eq. (\ref{Eq10}), we can simultaneously exploit multiple original features to learn an ensemble of base distance functions. The final distance is a weighted sum of these distance functions in Eq. (\ref{Eq.21}). Table \ref{Table3} shows the performance comparison of the proposed MVLDML+ method with two score-level fusion based methods: Ensem-XQDA and Ensem-MLAPG.

 By exploiting the first three visual GOG descriptors from the {\textrm{RGB}, \textrm{Lab}, \textrm{HSV}} color spaces, MVLDML+ achieves the best performance 48.86\% at rank-1 on VIPeR. 
 It outperforms Ensem-XQDA and Ensem-MLAPG over 1.14\% and 1.39\%, respectively, at rank-1. On CUHK01, the improvements are more obvious. MVLDML+ surpasses Ensem-XQDA and Ensem-MLAPG by over 3\% at rank-1. On PRID450S, the improvements are 0.27\% and 2.04\% at rank-1 respectively.
 It indicates that MVLDML+ can explore the complementary of different visual features more effectively. By exploring the complementary of the three descriptors, MVLDML+ improves the rank-1 recognition rate of LDML+ with GOG$_\textrm{RGB}$ by 3.67\%, and that of LDML+ with GOG$_\textrm{HSV}$ by 8.86\%.
  
We also report the performance of MVLDML+ using all the four GOG features in Table \ref{Table3}. By employing all the four descriptors, the rank-1 accuracy of MVLDML+ has been improved from 48.86\% to 50.03\% on VIPeR, 60.73\% to 61.37\% on CUHK01, and 64.80\% to 66.71\% on PRID450S.
The results shown in Table \ref{Table3} demonstrate the effectiveness of the MVLDML+ method.

 \begin{table*}[htbp]
	\caption{Top-ranked average recognition rates (CMC@rank-r, \%) of LDML+ and four baseline methods on the \textbf{Market-1501} dataset with three different visual feature descriptors. A larger number indicates a better result. The best results are shown in boldface.}
		\begin{center}	
			\renewcommand\arraystretch{1.2}
		{\begin{tabular}{c|l  |c c c >{\columncolor{mygray} }c  ||c c c >{\columncolor{mygray} }c ||c c c >{\columncolor{mygray} }c} 
				\ChangeRT{0.7pt}		
				\multicolumn{2}{c|} {\multirow{2}{*}{Market-1501 (SQ)}}    &\multicolumn{4}{c||}{ GOG$_\textrm{RGB}$} &\multicolumn{4}{c||}{ GOG$_\textrm{Lab}$}  &\multicolumn{4}{c}{GOG$_\textrm{HSV}$}\\\cline{3-14}
				\multicolumn{2}{c|}{}& R=1 & R=5 & R=10 & mAP& R=1 & R=5 & R=10 &mAP & R=1 & R=5 & R=10 & mAP\\
				\hline
				&XQDA\cite{LOMOXQDA} &{43.29} & {65.38}& {73.81} & 23.26& {43.32} & {65.65} & {74.26} & {23.30}& {35.24} & {56.84} & {68.29} & {17.55} \\
				Baseline&	  MLAPG\cite{MLAPG}     & 46.26 & 68.56& 77.05 & 24.13 & 47.48& 69.15& 77.94& 24.96& 40.26&64.43& 73.63& 19.79 \\
				methods&	LDML$^{\sigma=1}$   & 41.24& 65.29& 73.78 & 21.52 & 35.30 & 57.69& 66.48 & 16.83& 34.62& 60.15& 69.83& 16.65\\
				&LDML & 47.83 & 70.55 & 78.36& {25.67}   & 48.55& 70.43 & 78.71& 25.60& 40.91& 65.08& 74.47& 20.39\\
				\hline
				Ours&	LDML+ & \textbf{52.08}& \textbf{73.93}& \textbf{81.68}& \textbf{29.10}  & \textbf{52.29}& \textbf{74.47} & \textbf{81.59}& \textbf{28.72}  & \textbf{45.49} & \textbf{69.21}  & \textbf{78.47}  & \textbf{23.06}  \\
				\hline\hline
				\multicolumn{2}{c|} {\multirow{2}{*}{Market-1501 (MQ)}}    &\multicolumn{4}{c||}{ GOG$_\textrm{RGB}$} &\multicolumn{4}{c||}{ GOG$_\textrm{Lab}$}  &\multicolumn{4}{c}{GOG$_\textrm{HSV}$}\\\cline{3-14}
				\multicolumn{2}{c|}{}& R=1 & R=5 & R=10 & mAP& R=1 & R=5 & R=10 & mAP& R=1 & R=5 & R=10 & mAP\\
				\hline
				&XQDA\cite{LOMOXQDA} &{51.60} & {72.92}& {81.18} & 29.48 & {51.01} & {77.62} & {80.79} & {29.32}& {44.45} & {67.90} & {76.57} & {23.39} \\
				Baseline&	  MLAPG\cite{MLAPG}     & 56.50& 77.55& 84.35 & 30.99& 57.48& 79.54& 86.19& 32.12& 51.54 & 75.45& 82.72& 26.45\\
				methods&	LDML$^{\sigma=1}$   & 52.76& 75.65& 82.84& 28.11& 43.79 & 66.51 & 74.82 & 21.23& 45.87& 71.62& 79.96& 22.79\\
				&LDML & 58.61 & 79.63& 86.19& 32.97& 58.28& 80.40 & 86.67 & 33.00& 52.08& 76.37& 83.73& 27.34\\
				\hline
				Ours&	LDML+ & \textbf{63.63}& \textbf{83.05}  & \textbf{88.33}& \textbf{37.16}  & \textbf{62.53}& \textbf{83.28}& \textbf{88.48}& \textbf{37.05}  & \textbf{57.60} & \textbf{79.16}  & \textbf{85.39}  & \textbf{31.05}  \\
				\ChangeRT{0.7pt}		
			\end{tabular}}
	\end{center}
	\label{Table7}
\end{table*}

\begin{table*}[htbp]
	\caption{Top-ranked average recognition rates (CMC@rank-r, \%) of MVLDML+ and two baseline methods on the \textbf{Market-1501} dataset. A larger number indicates a better result. The best results are shown in boldface.}
		\begin{center}	
			\renewcommand\arraystretch{1.2}
			{\begin{tabular}{l|l|c c c >{\columncolor{mygray} }c  ||c c c >{\columncolor{mygray} }c } 
				\ChangeRT{0.7pt}
				\multicolumn{2}{c|}{Market-1501} & \multicolumn{4}{c||}{ SQ}  &\multicolumn{4}{c}{MQ} \\
				\hline
				{ GOG Features} &{ Methods } & \textrm{  R=1} & \textrm{ R=5} &\textrm{R=10} & mAP  &\textrm{  R=1} & \textrm{ R=5} &\textrm{R=10} & mAP\\  				 		 	
				\hline	
				\multirow{3}{*}{\textrm{RGB},\textrm{Lab},\textrm{HSV}}
				& Ensem-XQDA& 45.58	&67.07	&75.45	&25.4	&53.15	&74.88	&82.21		&31.72\\
				& Ensem-MLAPG &51.34	&73.01	&80.94	&28.23	&60.96	&81.26	&87.41	&35.56\\  \cline{2-10}
				&  {MVLDML+} & \textbf{55.94} & \textbf{77.94} & \textbf{84.35}& \textbf{32.19}  & \textbf{66.81} & \textbf{85.33} & \textbf{90.23}& \textbf{40.63}\\ 			
				\hline\hline
				\multirow{3}{*}{\textrm{RGB},\textrm{Lab},\textrm{HSV},\textrm{nRnG}} 
				& Ensem-XQDA&47.3	&68.14	&75.67	&26.36	&55.20	&75.50	&82.72	&32.65\\
				& Ensem-MLAPG &51.99	&73.57	&81.03	&28.62	&61.25	&81.15	&87.02	&	35.92\\  \cline{2-10}
				&  {MVLDML+} &\textbf{58.22}	&\textbf{78.59}	&\textbf{84.98}	&\textbf{33.7}	&\textbf{68.38}	&\textbf{86.28}	&\textbf{90.59}	&\textbf{41.84}		\\ 
				\ChangeRT{0.7pt}				
			\end{tabular}}
	\end{center}
	\label{Table8}
\end{table*}	
\subsubsection{Comparison with State-of-the-art Results} 
In Table \ref{Table4}, we compare the performance of MVLDML+ with most recent state-of-the-art results on VIPeR at different ranks. By using four GOG descriptors, our method achieves the best rank-1 recognition rate 50.0\%. Note that it is only slightly better than the result of GOG$_\textrm{Fusion}$+XQDA reported in \cite{GaussianReIDDescriptor}. That is because we use a different split of dataset with \cite{GaussianReIDDescriptor}. Our split is the same as that in MLAPG\cite{MLAPG}, where the dataset is randomly split with a fixed random seed ($\mathrm{rng(0)}$ on Matlab) for 10 times. If GOG$_\textrm{Fusion}$+XQDA is implemented using our split, it obtains 48.42\% at rank-1, 78.23\% at rank-5, and 87.63\% at rank-10, which is clearly lower than our results.

We also report the results of LDNS \cite{zhang2016learning} in Table \ref{Table4}, a more recent metric learning method, using the GOG$_\textrm{Fusion}$ descriptor. It formulates a much stricter learning objective than the classic Fisher discriminative analysis (FDA) method. 
It aims to minimize the within-class scatter by collapsing all the images of the same person into a single point. This learning objective may be too rigorous to cope with the GOG$_\textrm{Fusion}$ descriptor in which multiple different feature descriptors are fused, thus resulting inferior performance than MVLDML+ and XQDA.

The third best result in Table \ref{Table4} is obtained with a multi-channel parts-based convolutional neural network model \cite{cheng2016person} which has a very strong feature representation ability. Our result surpasses the third best result by 2.2\% at rank-1. 

Table \ref{Table5} compares the top-ranked recognition rate of MVLDML+ with most recent state-of-the-art results on CUHK01. We can observe that, owing to the effectiveness of locally adaptive decision rule, MVLDML+  achieves a state-of-the-art rank-1 accuracy 61.4\%, improving the result of GOG$_\textrm{Fusion}$+XQDA by 3.6\%.

Table \ref{Table6} compares the top-ranked recognition rates of MVLDML+ with several state-of-the-art results on PRID450S. It is observed that our method achieves the second best rank-1 recognition rate on PRID450S, which is comparable to deep network based result \cite{wu2016enhanced}. It can be mainly owed to the effectiveness of the locally adaptive decision rule and the multi-view learning strategy.

Note that, we only exploit four GOG color descriptors in this work. MVLDML+ can achieve a better performance by exploiting more visual features (e.g. WHOS \cite{WhosDescriptors}).

\subsection{Experiments on Market-1501}
{In this subsection, we evaluate the effectiveness of the proposed LDML+ and MVLDML+ methods on the large dataset, Market-1501 \cite{ZhengLiangReIDBenchmark}. The training set contains 12936 bounding boxes of 750 identities. The testing set contains 19732 bounding boxes of 751 identities. Compared to the above three small size datasets, Market-1501 has more complex intra-class and inter-class variations. To formulate a cross-camera setting, we split the training set into a probe set and a gallery set, where both the probe and gallery sets have 4914 training samples. 9828 training samples are finally used. 46518 positive pairs and over 20 million negative pairs are generated. 

Table VII and Table VIII show the performance evaluation of the LDML+ and MVLDML+ methods respectively. Table IX compares the top-ranked recognition rates and mAP scores of MVLDML+ with several state-of-the-art results on Market-1501. Both the single-query and multi-query results are provided.

\subsubsection{Evaluation of LDML+} 
As shown in Table VII, the proposed LDML+ method performs much better than the baseline methods on Market-1501, one of the largest datasets. In the single-query (SQ) setting, LDML+ obtains 52.08\%, 52.29\%, and 45.49\% at rank-1, respectively, and 29.10\%, 28.72\% and 23.06\% in terms of mAP, respectively, using the three GOG features. Our method beats the XQDA method, which performs very well on small size datasets, by a large margin. Using the GOG$_\textrm{RGB}$ feature, LDML+ surpasses XQDA by 8.79\% at rank-1 and 5.84\% in terms of mAP. Similar improvements can also be obtained using the other two features. It is mainly due to that the Gaussian assumption of XQDA does not hold any more on such a large dataset with much more complex intra-class and inter-class variations. 

In the SQ setting,  using the three features respectively, LDML+ improves the rank-1 recognition rates of its counterpart, LDML, by 4.25\%, 3.74\%, and 4.58\%, and the mAP scores by 3.43\%, 3.12\%, and 2.67\%. It reflects that the locally adaptive decision rule performs much better than global decision rule at the large dataset. It can better characterize the similarity relationship between a pair of person samples and guide the gradient descending toward the right direction, thus resulting a more reliable metric. 

In the multi-query (MQ) setting, using the three features respectively, the rank-1 recognition rates of LDML+ have been improved from 52.08\% to 63.63\%, 52.29\% to 62.53\%, and 45.49\% to 57.60\%.  Significant improvements of LDML+ on the baseline methods can also be observed in the MQ setting.

Besides, we find that the rank of metric M in LDML+ drops faster than that in the two global decision threshold based methods, LDML and MLAPG, on Market-1501 dataset, thus resulting a lower-rank yet more discriminative metric. 

The improvements shown in Table VII have clearly demonstrated the effectiveness of LDML+.

\begin{table}[tbp]
	\caption{Performance Comparison (CMC@rank-R and mAP, \%) on the \textbf{Market-1501} dataset. A larger number indicates a better result. Both single-query (SQ) and multi-query (MQ) evaluation results are presented respectively. }
	\begin{center}
		\renewcommand\arraystretch{1.2}
		{\begin{tabular}{l|c|c >{\columncolor{mygray} }c  ||c  >{\columncolor{mygray} }c } 
				\ChangeRT{0.7pt}
				\multicolumn{2}{c|} {\multirow{2}{*}{Market 1501} } &\multicolumn{2}{c||}{SQ} &\multicolumn{2}{c}{MQ}  \\\cline{3-6}
				\multicolumn{2}{c|}{}& R=1  & mAP & R=1  & mAP \\
				\hline
				MVLDML+ & Ours & 58.22 & 33.70  &68.38  &41.84 \\
				MVLDML+ (Re)& Ours & 64.82 & \textbf{48.01} &74.58 &\textbf{56.45}\\
				BoW+KISSME\cite{ZhengLiangReIDBenchmark}& ICCV 2015 & 44.42 & 20.76& -& - \\
				LOMO+XQDA\cite{LOMOXQDA}& CVPR 2015 & 43.80 & 22.20& -& - \\
				WARCA\cite{jose2016scalable}& ECCV 2016 & 45.16 &- & -& - \\
				LOMO+LDNS\cite{zhang2016learning} & CVPR2016&55.43 & 29.87& 67.96& 41.89\\
				SLSC\cite{ChenDaPeng_similarityLearning} & CVPR2016 & 51.90 &  26.35& - &  - \\
				TMA\cite{martinel2016temporal} &ECCV 2016 &47.92 & 22.31& -& -\\
				AttentionNet\cite{liu2017end}& TIP 2017  & 48.24&24.43& -& -\\
				DeepAttribute\cite{su2016deep} & ECCV 2016 & 39.40&19.60& 49.00&25.80\\
				MSTripletCNN\cite{liu2016multi} & MM 2016 & 45.10 &- & 55.40&-\\
				DeepEmbedd\cite{ustinova2016learning} & NIPS 2016 & 59.47& -& -& -\\
				SiameseLSTM\cite{varior2016siamese}&ECCV 2016&- & -& 61.60&35.30\\
				ContrastiveLoss\cite{varior2016gated}&ECCV 2016& 62.32& 36.23& 72.92& 45.39\\
				GatedSiamese\cite{varior2016gated}&ECCV 2016& \textbf{65.88}& 39.55& \textbf{76.04}& 48.45\\
				\ChangeRT{0.7pt}	  			
		\end{tabular}}
	\end{center}
	\label{Table9}
	\vspace{-0.050in}
\end{table}
\subsubsection{Evaluation of MVLDML+} 
As shown in Table VIII, by employing the three GOG features as input, MVLDML+ obtains 55.94\% rank-1 recognition rate and 32.19\% mAP score in SQ setting, and 66.81\% rank-1 recognition rate and 40.63\% mAP score in MQ setting. Comparing the results of MVLDML+ in Table VIII with the results of LDML in Table VII, we can find that MVLDML+ improves the performance of LDML+ by a large margin. It reveals that the complementary of multiple feature descriptors has been well exploited.

Table VIII compares the performance of MVLDML+ with Ensem-XQDA and Ensem-MLAPG in both the SQ and MQ settings. By only using three GOG features, MVLDML+ improves the performance of Ensem-XQDA by over 10\% at rank-1 in SQ setting and 13\% at rank-1 in MQ setting. It also beats Ensem-MLAPG by over 4\% at rank-1 in SQ setting and 5\% at rank-1 in MQ setting. Similar improvements can also be observed when all the four GOG features have been employed in MVLDML+.
The results shown in Table VIII demonstrate the effectiveness of MVLDML+.

\subsubsection{Comparison with State-of-the-art Results} 	
In Table IX, we compare the performance of MVLDML+ with most recent state-of-the-art results on Market-1501. Both the rank-1 recognition rate and mAP score are used as the evaluation metrics. We also apply a re-ranking strategy \cite{Zhong2017CVPR} to boost the performance of MVLDML+ in testing stage, which is termed as MVLDML+ (Re) in Table IX.

It should be noted that most recent results in Table IX are obtained by training deep neural networks on Market-1501, which can yield much stronger features. For example, Varior et al. \cite{varior2016gated} presents a gated Siamese architecture that yields the best rank-1 recognition rate in both SQ and MQ settings in Table IX. However, our proposed method uses the hand-crafted GOG color descriptors as input to learn a low-rank and discriminative metric, obtaining 58.22\% rank-1 recognition rate in SQ setting and 68.38\% rank-1 recognition rate in MQ setting. Although it is lower than the results in \cite{varior2016gated}, it still performs better than most listed results in Table IX including multiple results of deep models \cite{liu2017end,liu2016multi,ustinova2016learning,varior2016siamese}. By employing the re-ranking technique, our results can be significantly boosted. MVLDML+ (Re) yields the best mAP scores (48.01\% and 56.45\%) in both SQ and MQ settings. It obtains 64.82\% rank-1 recognition rate in SQ setting and 74.58\% rank-1 recognition rate in MQ setting, which is comparable to the results of \cite{varior2016gated}. }

\subsection{Analysis of the proposed method}
\subsubsection{On the parameter $\beta$}
\begin{figure}[tbp]
	\begin{center}
		\begin{minipage}[b]{0.5\textwidth}
			\includegraphics[width=1.7in]{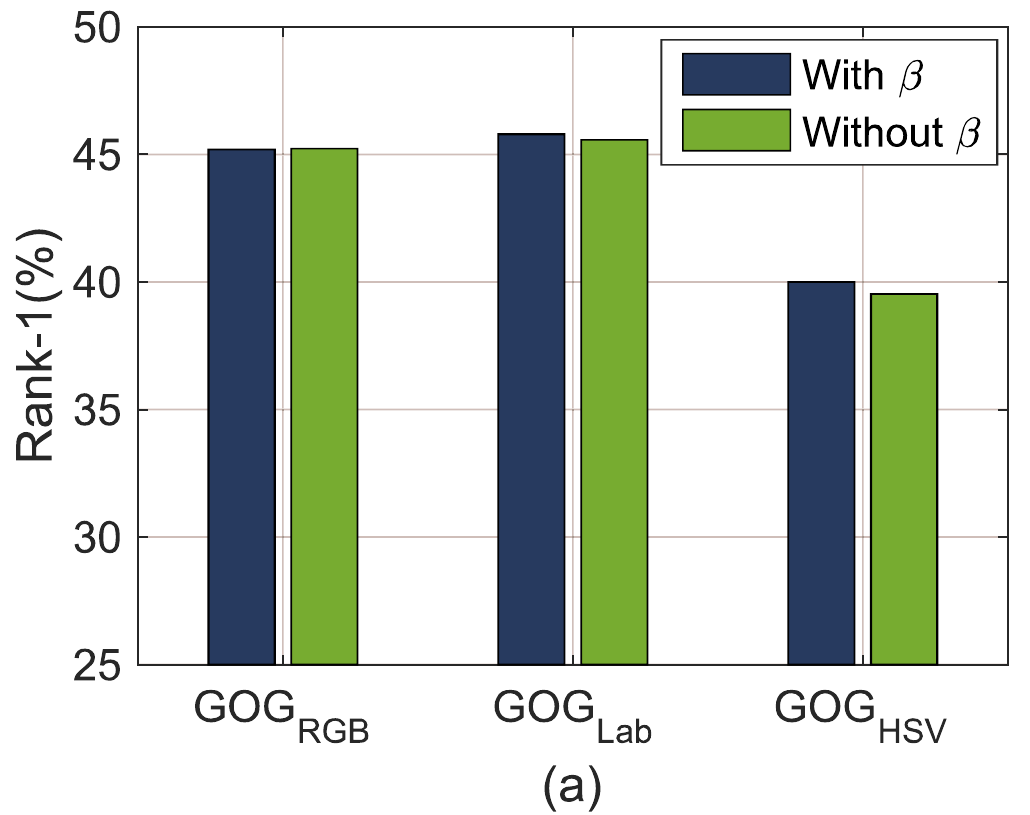}
			\includegraphics[width=1.7in]{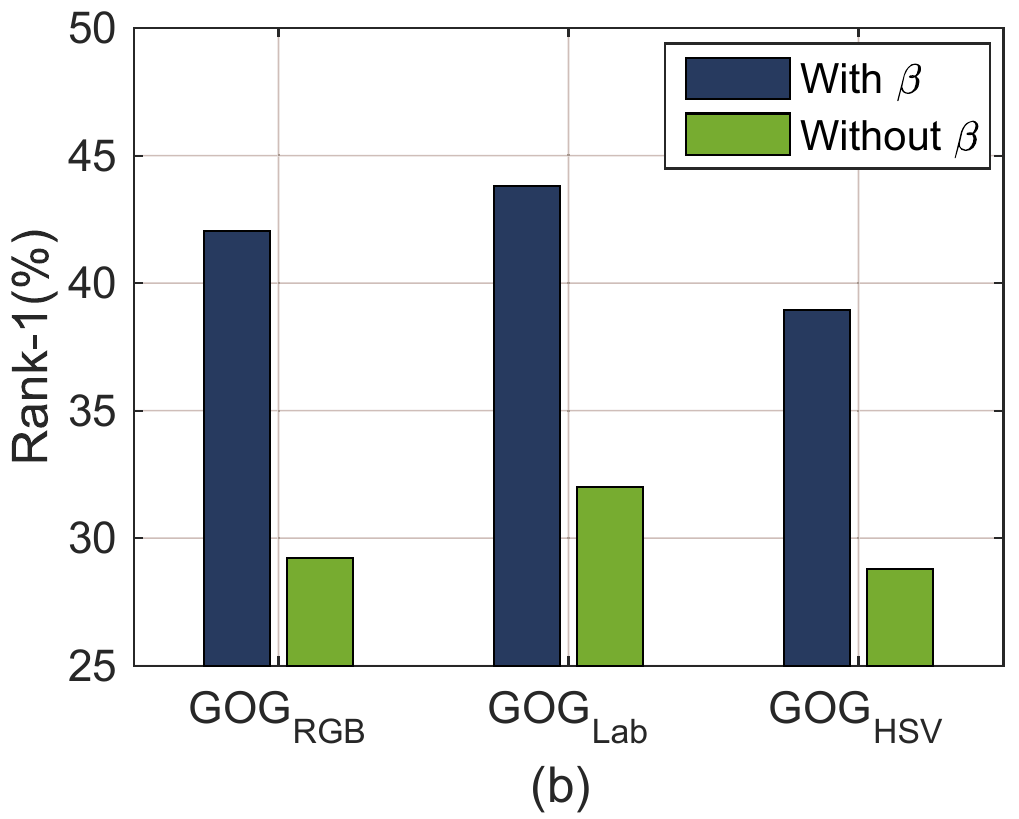} 
		\end{minipage}
	\end{center}
	\vspace{-0.1500in}
	\caption{Performance comparison of LDML+ with and without the parameter $\beta$ on the VIPeR dataset. (a) $\{\mathbf{x}^*\}$ are represented by the privileged features; (b) $\{\mathbf{x}^*\}$ are represented by randomly-generated vectors. }
	\vspace{-0.10in}
	\label{figure2}
\end{figure}
{We introduce in Eq. (\ref{Eq5}) a scale parameter $\beta=\frac{mean(\mathbf{D}_\mathbf{M})
}{mean(\mathbf{D}_\mathbf{P})}$ that is expected to globally bridge the gap between the privileged features $\{\mathbf{x}^*\}$ and original features $\{\mathbf{x}\}$. The reason for that is, in some case, $\{\mathbf{x}^*\}$ may have a significantly different distribution from $\{\mathbf{x}\}$, especially when the privileged information is partially or even totally wrong. Here, the scale parameter $\beta$ can smooth the distance $\mathbf{D}_\mathbf{P}$ in privileged space and is helpful for searching a suitable step-size at the beginning of the optimization. In this subsection, we compare the performance of LDML+ with and without $\beta$ in two cases. One is representing $\{\mathbf{x}^*\}$ with the privileged features by the setting in Section \ref{5.1}. The other case is replacing the privileged features with randomly-generated vectors. The performance comparison is shown in Fig. \ref{figure2}, where the VIPeR dataset is used as an example.

In the first case shown in Fig. \ref{figure2}(a), the improvements of LDML+ with $\beta$ on that without $\beta$ is not obvious. However, in the second case shown in Fig. \ref{figure2}(b), where the privileged information is totally wrong, LDML+ with $\beta$ significantly surpasses LDML+ without $\beta$. It shows that in some extreme case where privileged information is totally wrong, the scale parameter $\beta$ does help. With $\beta$, LDML+ can still yield satisfactory results, although lower than the results in Fig. \ref{figure2}(a). Without $\beta$, the gradient of $\mathbf{M}$ would descend toward a totally wrong direction, thus resulting a bad performance.

The results in Fig. \ref{figure2} clearly demonstrates the effectiveness of the scale parameter $\beta$.}

\subsubsection{On the parameter $\lambda$}

\begin{figure}[tbp]
	\begin{center}
		\begin{minipage}[b]{0.5\textwidth}
			\includegraphics[width=1.65in]{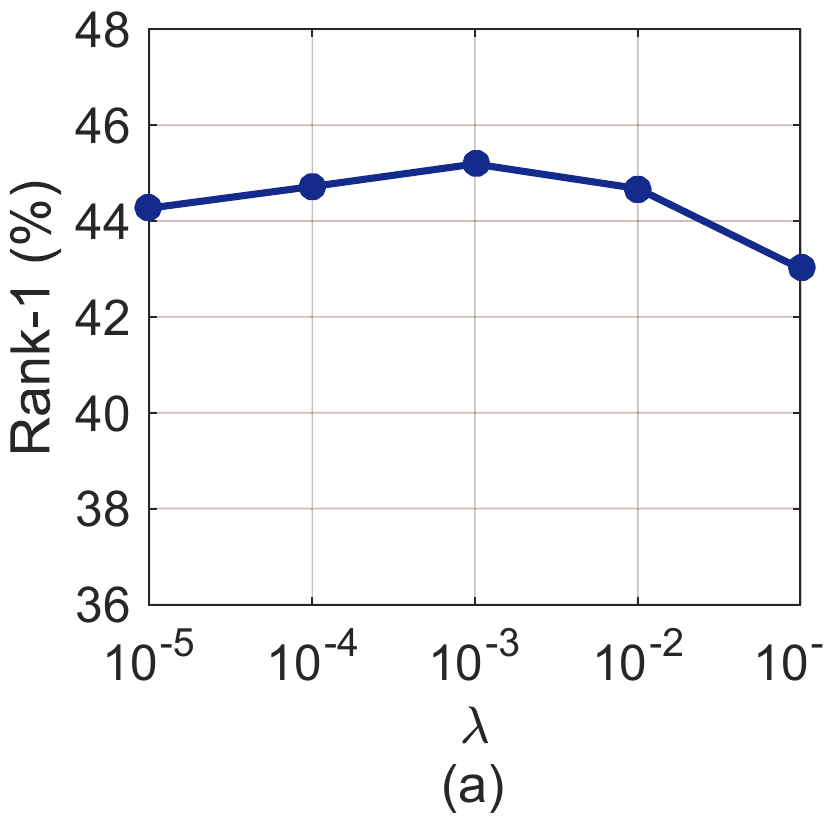}
			\includegraphics[width=1.8in]{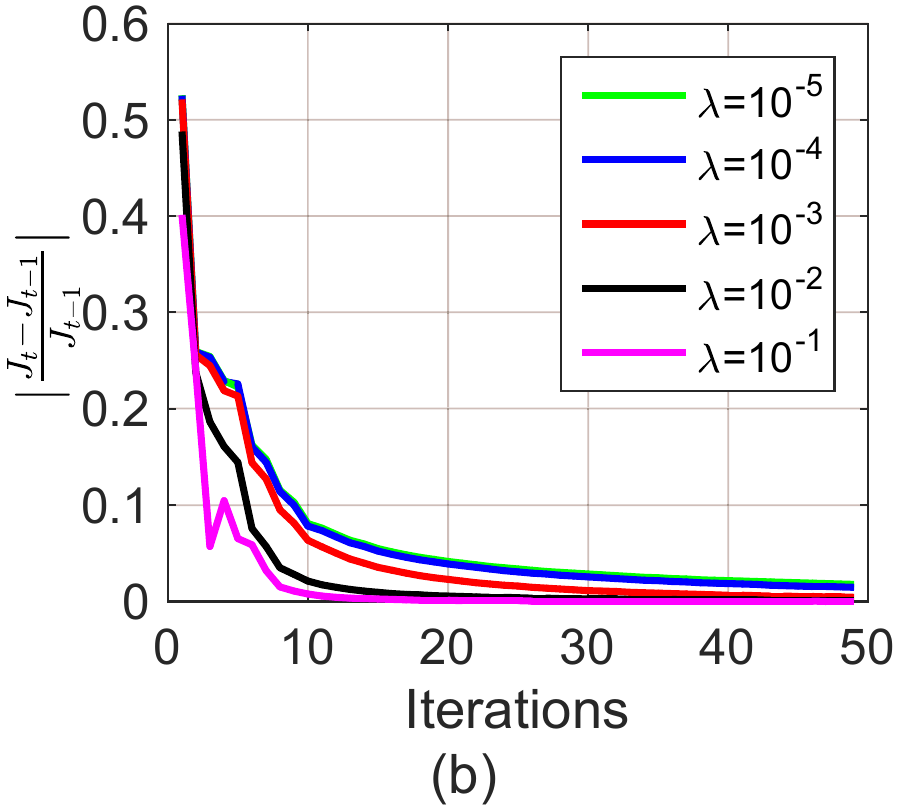} 
		\end{minipage}
	\end{center}
	\vspace{-0.1500in}
	\caption{Effects of the parameter $\lambda$ on the performance and convergence of LDML+. (a) Rank-1 recognition rates (b) Convergence curves.}
	\vspace{-0.10in}
	\label{figure3}
\end{figure}

The parameter $\lambda$ modulates the effect of the regularization term $\mathcal{R}(\mathbf{P})=\Vert \mathbf{P} \Vert^2_F/{d^*}$. If $\lambda$ is too small, the metric $\mathbf{P}$ will have higher degree of freedom, which may result in slow convergence. While, a large $\lambda$ may degrade the performance of our method because of premature convergence.
In this subsection, we investigate the effects of $\lambda$ on the performance and convergence of the proposed LDML+ method. Here, the VIPeR dataset is used as an example. We use the GOG$_\textrm{RGB}$ descriptor as original feature representation.
We illustrate the changes of rank-1 recognition rate of the LDML+ method in Fig. \ref{figure3}(a) by varying $\lambda$ from $10^{-5}$ to $10^{-1}$. To analyze of the effect of $\lambda$ on the convergence, we use $\vert\frac{J_t-J_{t-1}}{J_{t-1}}\vert$ as the objective value and observe the changes of objective value with different settings of $\lambda$ in Fig. \ref{figure3}(b).

It can be seen from Fig. \ref{figure3}(a) that the rank-1 recognition rate of LDML+ changes little when $10^{-5}\leq\lambda \leq 10^{-2}$. When $\lambda>10^{-2}$, the performance drops fast. It shows that our method is sensitive to a large $\lambda$. As shown in Fig. \ref{figure3}(b),  the larger the parameter $\lambda$ is, the faster the algorithm converges. When $\lambda=10^{-1}$, the algorithm converges in less than 20 iterations but with the lowest rank-1 recognition rate, since the algorithm has fallen in a bad local optima.

We empirically set $\lambda=10^{-3}$ as a trade-off between the performance and the convergence speed on the three small size datasets. For the large dataset, Market-1501, we set $\lambda$ to $10^{-4}$.
\subsubsection{Performance comparison at varying PCA dimensions}
\begin{figure}[tbp]
	\begin{center}
		\includegraphics[width=2.8in]{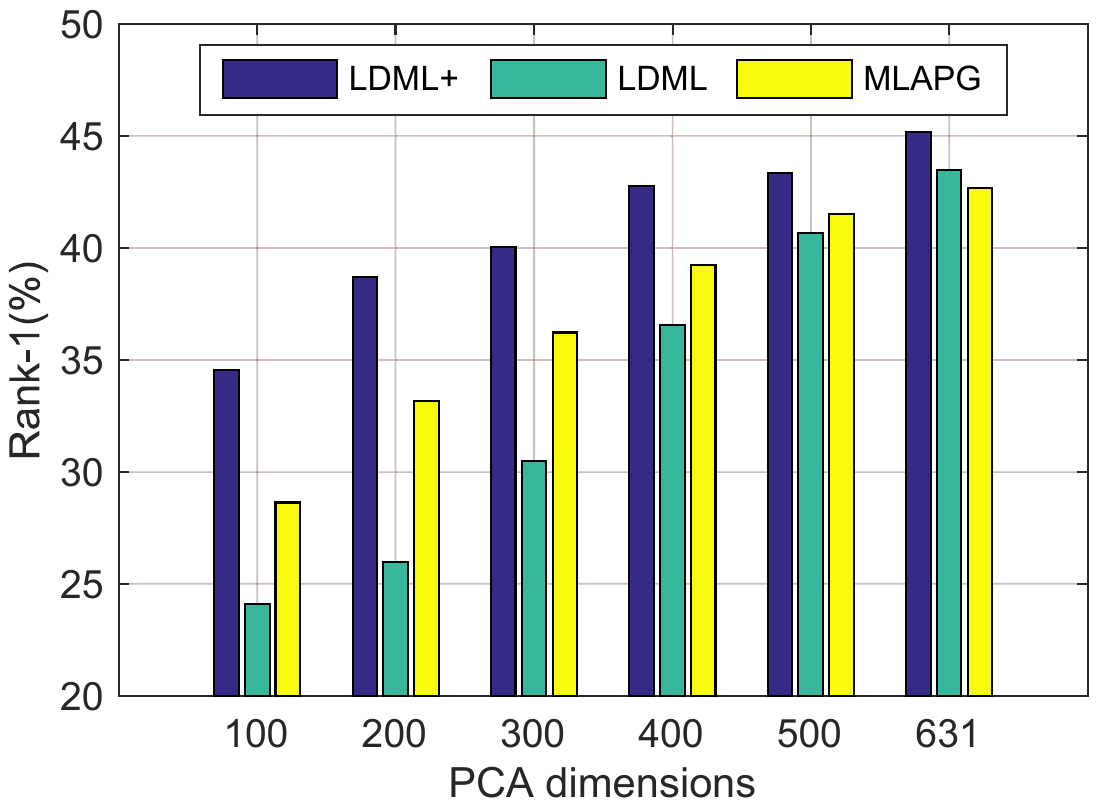}
	\end{center}
	\vspace{-0.1500in}
	\caption{Performance comparison of LDML+ with LDML and MLAPG at varying PCA dimensions.}
	\vspace{-0.10in}
	\label{figure4}
\end{figure}

{PCA is applied for dimension reduction in this work but almost all energy is retained. In this subsection, we compare the performance of LDML+ with LDML and MLAPG at varying PCA dimensions. The VIPeR dataset is used as an example and the GOG$_\textrm{RGB}$ feature is employed.
Fig. \ref{figure4} compares the rank-1 recognition rates of LDML+ with two baseline methods at varying PCA dimensions $d\in\{100, 200, 300, 400, 500, 631\}$. Here, 631 is the full PCA dimension on VIPeR.

As shown in Fig. \ref{figure4}, when $d$ is high (e.g. 631) where more energy is retained, the improvement of LDML+ on the two baseline methods is small. Nearly 2\% improvement at rank-1 is observed when $d=500$. With the decreasing of the dimension $d$, the advantage becomes more obvious.
When the dimension is dropped to 100, LDML+ yields remarkable improvements on the two global decision threshold based methods, over 10\% on LDML and 5\% on MLAPG at rank-1. It is mainly because that given a much lower-dimensional representation, LDML and MLAPG may be more prone to overfitting on training set. While, benefiting from the locally adaptive decision rule built with privileged features, LDML+ can yield a metric with higher generalization capacity on testing set. Besides, high-dimension original features just like a \textit{student} with strong learning ability. It can already learn a good metric without the help of \textit{teacher} (privileged features).
While, the low-dimensional features just like a \textit{student} with poor learning ability, it can gain more help from the \textit{teacher}.}

\subsubsection{The effects of the privileged metric $\mathbf{P}$}
{
In this subsection, we investigate the effects of the privileged metric $\mathbf{P}$ on the Market-1501 dataset in SQ setting. GOG$_\textrm{RGB}$ feature is used as an example. Fig. \ref{figure5} presents normalized distance histograms of positive training pairs and negative training pairs on both the original feature space and the privileged feature space before/after metric learning. 

\begin{figure}[tbp]
	\begin{center}
		\begin{minipage}[b]{0.52\textwidth}
			\includegraphics[width=1.65in]{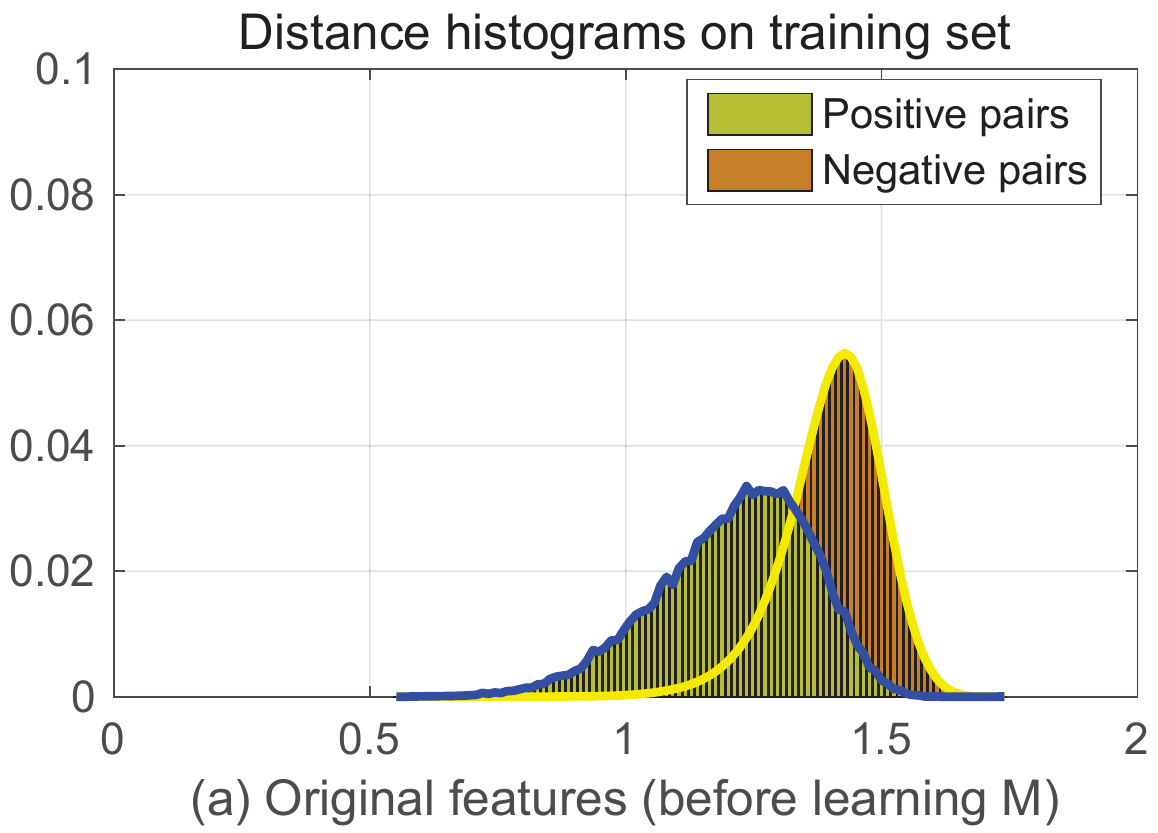}
			\includegraphics[width=1.65in]{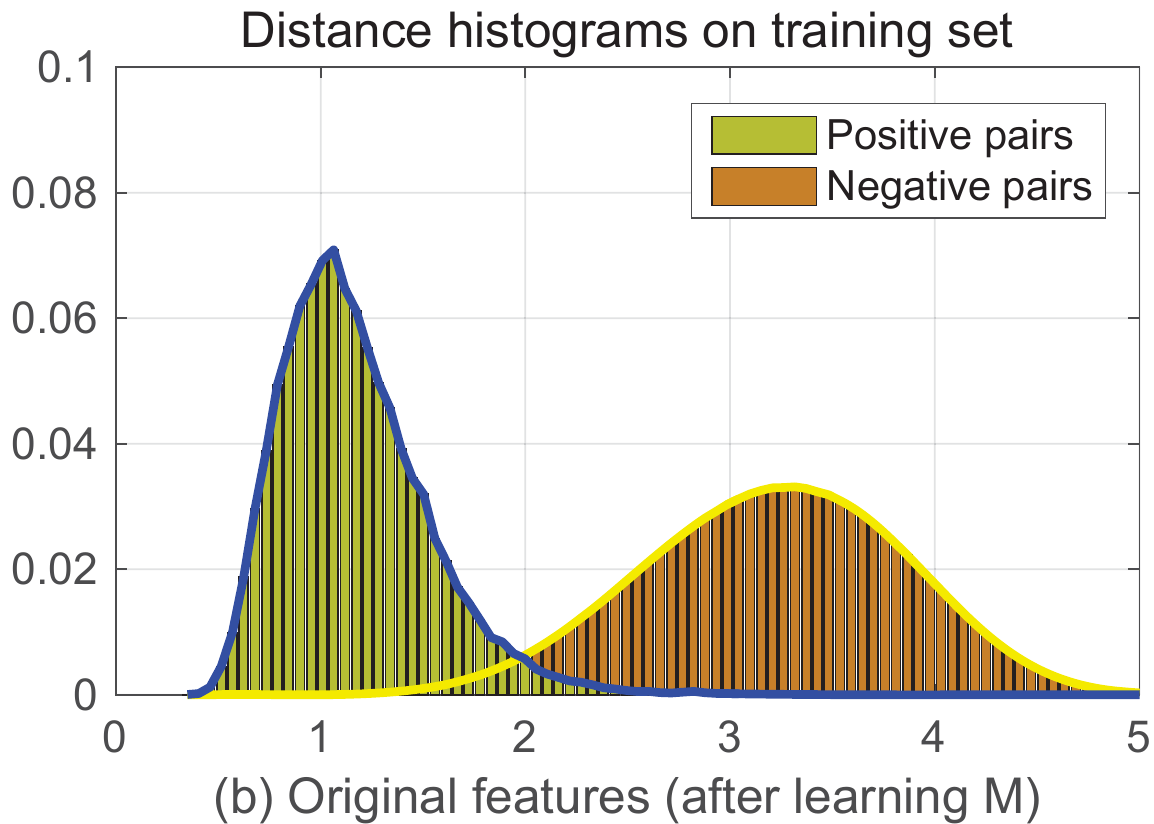} 
		\end{minipage}
		\begin{minipage}[b]{0.5\textwidth}
			\includegraphics[width=1.65in]{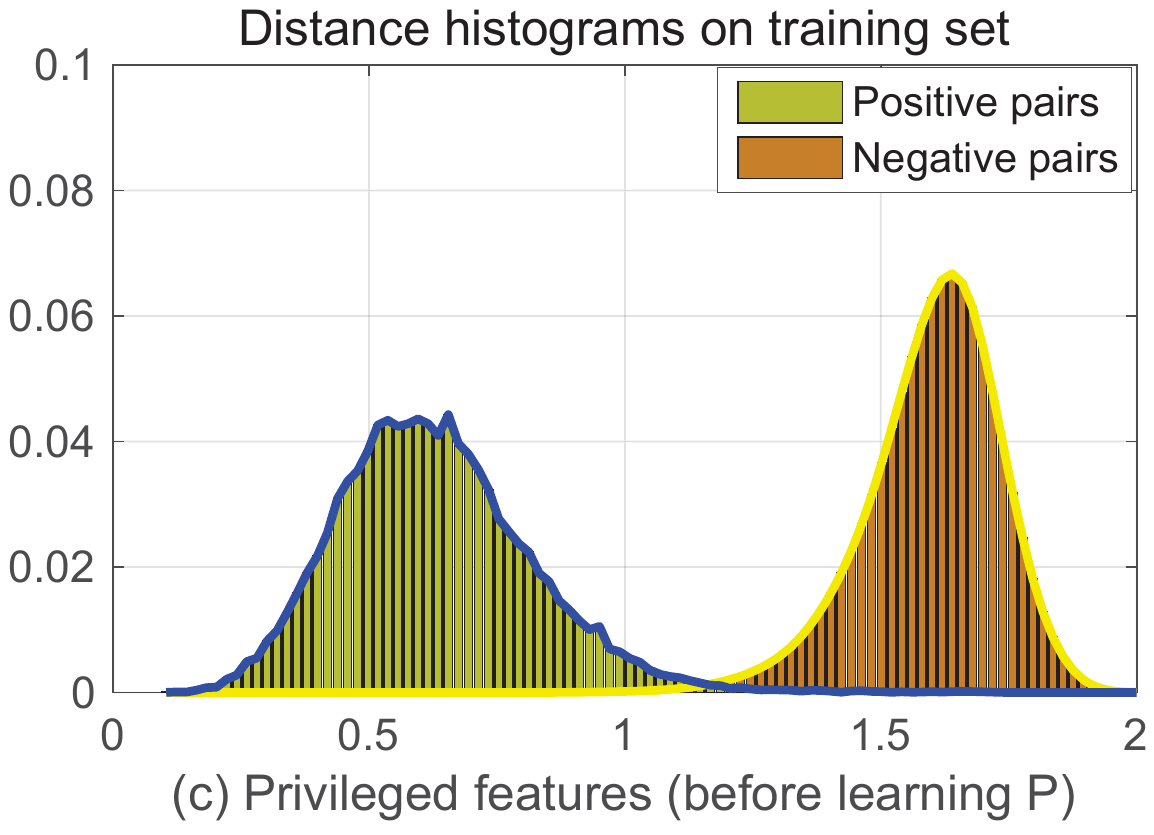}
			\includegraphics[width=1.7in]{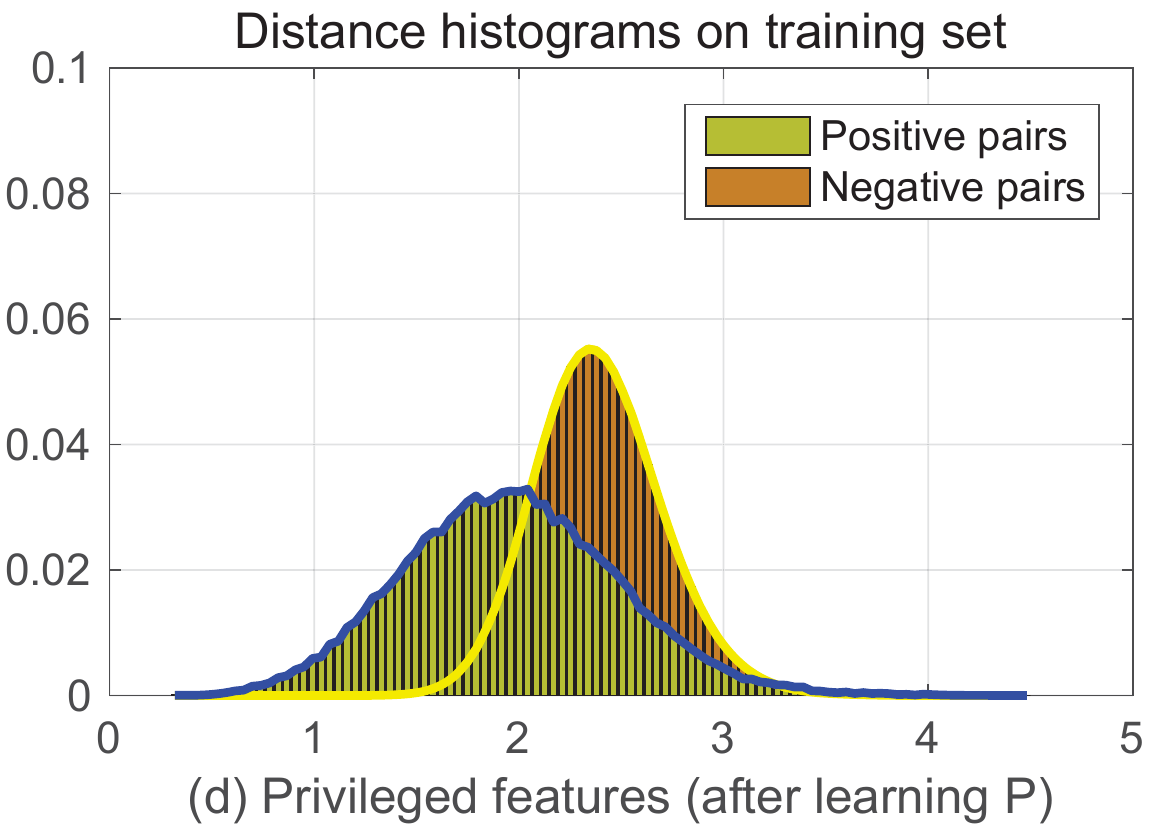} 
		\end{minipage}
	\end{center}
	\vspace{-0.1500in}
	\caption{{Normalized distance histograms on the training set of Market-1501 before/after metric learning.  (a) On the original feature space before learning $\mathbf{M}$; (b) On the original feature space after learning $\mathbf{M}$; (c) On the privileged feature space before learning $\mathbf{P}$; (d) On the privileged feature space after learning $\mathbf{P}$.}
	}
	\vspace{-0.10in}
	\label{figure5}
\end{figure}

In this work, the metric $\mathbf{P}$ is always associated with the distance \(d_\mathbf{P}^2(\mathbf{x}_i^*,\mathbf{z}_i^*)\) that functions as a local decision threshold to guide the learning of the target metric $\mathbf{M}$. During the training stage, given a positive pair, the distance \(d_\mathbf{M}^2(\mathbf{x}_i,\mathbf{z}_i)\) is expected to be smaller than \(\beta d_\mathbf{P}^2(\mathbf{x}_i^*,\mathbf{z}_i^*)\); given a negative pair,  the distance \(d_\mathbf{M}^2(\mathbf{x}_i,\mathbf{z}_i)\) is expected to be larger than \(\beta d_\mathbf{P}^2(\mathbf{x}_i^*,\mathbf{z}_i^*)\). This expectation has been clearly illustrated by Fig. \ref{figure5} (b) and (d). The reason of jointly learning metric $\mathbf{P}$ with $\mathbf{M}$ is that it may be too rigorous to directly use the Euclidean distances in the privileged feature space as the decision threshold for metric learning, due to the significant difference between the privileged feature distribution and the original feature distribution, which can be observed in Fig. \ref{figure5} (a) and (c). Therefore, the privileged distance $d_\mathbf{P}^2(\mathbf{x}_i^*,\mathbf{z}_i^*)$ is constantly adapted for guiding the learning of $\mathbf{M}$ during the training stage. As shown in Fig. \ref{figure5} (a) and (b), benefiting from jointly learning metric $\mathbf{P}$ with $\mathbf{M}$, almost all the positive pairs on the original feature space has been distinguished from the negative pairs.
	
Fig. \ref{figure6} compares the performance of LDML+ with/without learning $\mathbf{P}$ in terms of rank-1 recognition rate and mAP score. Here, the performance of \textit{LDML+ without learning $\mathbf{P}$} is obtained by directly employing the Euclidean distance on privileged space as the decision threshold. We also use Euclidean distance (without training) of original features for person matching on testing set and employ its performance as a baseline in Fig. \ref{figure6}.
As shown in Fig. \ref{figure6}, if directly using Euclidean distance for person matching, we can only achieve 22.74\% rank-1 recognition rate and 8.07\% mAP score.
By learning a Mahalanobis distance on training set, the rank-1 recognition rate and mAP score have been significantly improved. 
By jointly learning $\mathbf{P}$ with $\mathbf{M}$, we obtain 52.08\% rank-1 recognition rate and 29.10\% mAP score, which improves the performance of that without learning $\mathbf{P}$ by over 13\% rank-1 recognition rate and 9\% mAP score.

Here, the metric $\mathbf{P}$ bridges the original feature and privileged feature, enabling the knowledge of \textit{teacher} to be smoothly transferred from privileged space to original space where \textit{student} makes a decision.}
\begin{figure}[tbp]
	\begin{center}
		\includegraphics[width=3in]{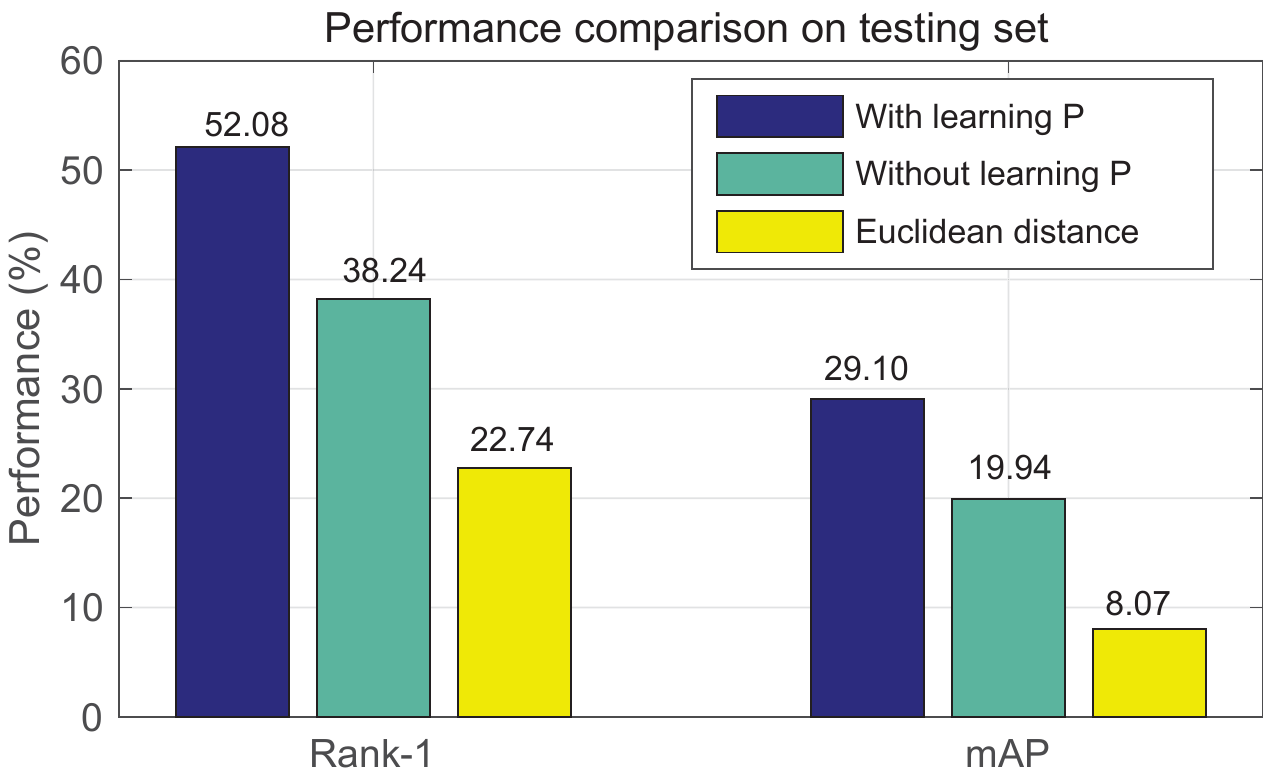} 
	\end{center}
	\vspace{-0.1500in}
	\caption{Performance comparison of LDML+ with/without learning the privileged metric $\mathbf{P}$.}
	\vspace{-0.10in}
	\label{figure6}
\end{figure}
\section{Conclusion}
In this paper, we develop a logistic discriminant metric learning approach for cross-view person re-ID. It exploits privileged information to build a locally adaptive decision rule which can cope well with complex inter-class and intra-class variations. Besides, the proposed approach is extended to a multi-view setting, which explores the complementation of multiple different visual representations effectively. In addition, an effective iterative optimization strategy is introduced to solve the proposed method. Extensive experimental evaluations and analyses on multiple challenging datasets have demonstrated the effectiveness of the proposed work.

\bibliographystyle{./IEEEtran}
\bibliography{mybib}

\begin{IEEEbiography}[{\includegraphics[width=1in,clip,keepaspectratio]{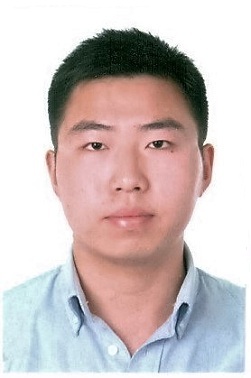}}]{Xun Yang}
	is pursuing the Ph.D. degree with the School of Computer and Information Engineering, Hefei University of Technology, China. 
	He was a visiting research student with the Centre for Quantum Computation \& Intelligent Systems in the Faculty of Engineering and Information Technology, University of Technology Sydney from Oct. 2015 to Oct. 2017. His research interests include person re-identification, multimedia content analysis, computer vision and pattern recognition.
\end{IEEEbiography}

\begin{IEEEbiography}[{\includegraphics[width=1in,clip,keepaspectratio]{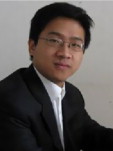}}]{Meng Wang} (M'09) is currently a professor at the Hefei University of Technology, China. He received his B.E. degree and Ph.D. degree in the Special Class for the Gifted Young and the Department of Electronic Engineering and Information Science from the University of Science and Technology of China (USTC), Hefei, China, in 2003 and 2008, respectively. His current research interests include multimedia content analysis, computer vision, and pattern recognition. He has authored more than 200 book chapters, journal and conference papers in these areas. He is the recipient of the ACM SIGMM Rising Star Award 2014. He is an associate editor of IEEE Transactions on Knowledge and Data Engineering (IEEE TKDE), IEEE Transactions on Circuits and Systems for Video Technology (IEEE TCSVT), and IEEE Transactions on Neural Networks and Learning Systems (IEEE TNNLS).
\end{IEEEbiography}


\begin{IEEEbiography}[{\includegraphics[width=1in,clip,keepaspectratio]{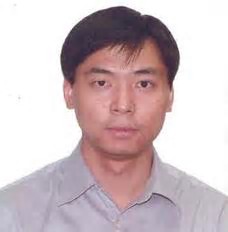}}]{Dacheng Tao} (F'15) is is currently a Professor of Computer Science in the School of Information Technologies at The University of Sydney. He mainly applies statistics and mathematics to artificial intelligence and data science. His research interests spread across computer vision, data science, image processing, machine learning, and video surveillance. His research results have expounded in one monograph and over 200 publications at prestigious journals and prominent conferences, such as the IEEE T-PAMI, T-NNLS, T-IP, JMLR, IJCV, NIPS, ICML, CVPR, ICCV, ECCV, AISTATS, ICDM, and ACM SIGKDD, with several best paper awards, such as the best theory/algorithm paper runner up award in the IEEE ICDM07, the best student paper award in the IEEE ICDM13, and the 2014 ICDM 10-year highest-impact paper award. He is a fellow of the IEEE, OSA, IAPR, and SPIE. He received the 2015 Australian Scopus-Eureka Prize, the 2015 ACS Gold Disruptor Award, and the 2015 UTS Vice-Chancellors Medal for Exceptional Research.

\end{IEEEbiography}
\end{document}